\documentclass[10pt,twocolumn,letterpaper]{article}

\usepackage{cvpr}      
\usepackage[accsupp]{axessibility}

\def\R{\mathbb{R}}

\def\e{\varepsilon}

\usepackage{threeparttable}
\usepackage{algorithm}
\usepackage{algorithmic}

\usepackage[most]{tcolorbox}
\newcounter{algcounter}
\usepackage{xcolor}

\usepackage{stackengine}
\newcommand\barbelow[1]{\stackunder[1.2pt]{$#1$}{\rule{.8ex}{.075ex}}}

\def\our{SHIELD}

\usepackage{tikz}
\usepackage{amsmath, amsfonts, amssymb, amsthm}

\theoremstyle{definition}
\newtheorem{definition}{Definition}[section]
\newtheorem{theorem}{Theorem}[section]
\usepackage{subcaption}
\usepackage{graphicx}
\usepackage{multirow}
\usepackage{booktabs}
\usepackage[utf8]{inputenc}
\usepackage[T1]{fontenc}

\definecolor{cvprblue}{rgb}{0.21,0.49,0.74}
\usepackage[pagebackref,breaklinks,colorlinks,allcolors=cvprblue]{hyperref}

\def\paperID{} 
\def\confName{CVPR}
\def\confYear{2026}

\title{\our{}: Secure Hypernetworks for Incremental Expansion Learning Defense}

\author{
Patryk~Krukowski\textsuperscript{1,2,3} \quad
Łukasz~Gorczyca\textsuperscript{1,2} \quad
Piotr~Helm\textsuperscript{1,2} \quad
Kamil~Ksi\k{a}\.{z}ek\textsuperscript{1} \\
Przemysław~Spurek\textsuperscript{1,4} \\
\vspace{0.2cm}\\
\textsuperscript{1}Jagiellonian University, Faculty of Mathematics and Computer Science \\
\textsuperscript{2}Jagiellonian University, Doctoral School of Exact and Natural Sciences \\
\textsuperscript{3}Akces NCBR \qquad \textsuperscript{4} IDEAS Research Institute \\
{\tt\small patryk.krukowski@doctoral.uj.edu.pl} \\
{\tt\small \{lukasz.gorczyca, piotr.helm, kamil.ksiazek, przemyslaw.spurek\}@uj.edu.pl}
}

\begin{document}
\maketitle
\begin{abstract}
Continual learning under adversarial conditions remains an open problem, as existing methods often compromise either robustness, scalability, or both. We propose a novel framework that integrates Interval Bound Propagation (IBP) with a hypernetwork-based architecture to enable certifiably robust continual learning across sequential tasks. Our method, \our{}, generates task-specific model parameters via a shared hypernetwork conditioned solely on compact task embeddings, eliminating the need for replay buffers or full model copies and enabling efficient over time. To further enhance robustness, we introduce Interval MixUp, a novel training strategy that blends virtual examples represented as $\ell_{\infty}$ balls centered around MixUp points. Leveraging interval arithmetic, this technique guarantees certified robustness while mitigating the wrapping effect, resulting in smoother decision boundaries. We evaluate \our{} under strong white-box adversarial attacks, including PGD and AutoAttack, across multiple benchmarks. It consistently outperforms existing robust continual learning methods, achieving state-of-the-art average accuracy while maintaining both scalability and certification. These results represent a significant step toward practical and theoretically grounded continual learning in adversarial settings. The code is available at \url{https://github.com/pkrukowski1/SHIELD}.
\end{abstract}    
\section{Introduction}
\label{sec:intro}

Deep neural networks have achieved remarkable performance across a wide range of tasks, often surpassing human-level accuracy \cite{he2016deep}. However, their success hinges on static training settings that do not reflect how intelligent systems are expected to operate in the real world. In practice, models must continually adapt to changing conditions while retaining prior knowledge, a challenge known as continual learning \cite{mccloskey1989catastrophic,hsu2018re}. At the same time, these models remain highly vulnerable to adversarial perturbations; small, carefully crafted input changes that can cause confident but incorrect predictions \cite{szegedy2013intriguing, goodfellow2014explaining}. This vulnerability raises serious concerns in high-stakes applications such as autonomous driving, robotics, and healthcare, where both adaptability and robustness are essential for safety and trust.

Surprisingly, most research treats these challenges in isolation. This disconnect is striking, given that real-world AI systems must not only learn and adapt continuously but also resist adversarial manipulation over time. Consider an autonomous vehicle that incrementally learns to navigate unfamiliar roads or environments. As it learns on the go, it must remain robust against adversarial threats, such as deceptive road signs, tampered traffic signals, and sensor manipulations that aim to mislead it. Without such resilience, this adaptive learning could become a critical vulnerability instead of an asset.

\begin{figure*}[]
    \begin{center}
    \begin{tikzpicture}[scale=0.25]
    \node[inner sep=0pt] (russell) at (-5.0,0)
    {\includegraphics[trim={0,8cm, 0,5cm, 0,1cm, 0,8cm},clip,width=0.98\linewidth]{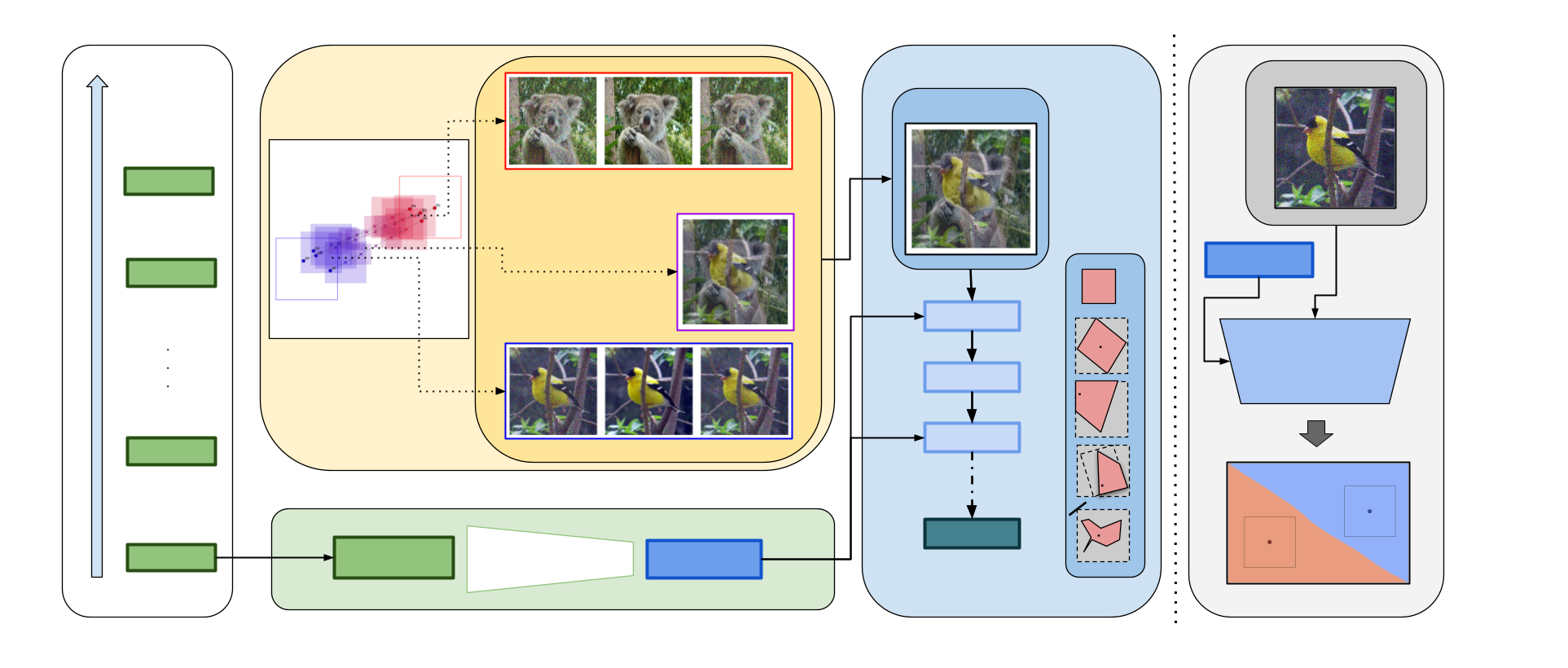} };
    
    \node[scale=0.7] at (-34,14.0) {Trainable embeddings};    

    \node[scale=0.6] at (-33,6.92) {$e_{T}$};
    \node[scale=0.6] at (-33,2.92) {$e_{T-2}$};
    \node[scale=0.6] at (-33,-5.2) {$e_{t+1}$};  
    \node[scale=0.6] at (-33,-10.0) {$e_{t}$};  
    
    \node[scale=0.7] at (-23,-10.0) {$e_{t}$};  
    \node[scale=0.65] at (-16,-10.0) {Hypernetwork};
    \node[scale=0.7] at (-9.0,-10.0) {$\mathcal{H}( \mathbf{e}_{t} ; \boldsymbol{\Phi} )$};

    \node[scale=0.7] at (-16,14.0) {Interval MixUp};
    \node[scale=0.5] at (-14.0,3.8) {$\lambda x_A + (1-\lambda) x_B$};
    \node[scale=0.4] at (-7.5, 6) {Interval MixUp Image};
    \node[scale=0.4] at (-7.2, 7.1) {$x_B + \left[-\e_2,\e_2\right]$};
    \node[scale=0.4] at (-11.5, 7.1) {$x_B$ (original)};
    \node[scale=0.4] at (-15.6, 7.1) {$x_B +\left[-\e_1,\e_1\right]$};

    \node[scale=0.4] at (-7.2,-5.1) {$x_A +\left[-\e_2,\e_2\right]$};
    \node[scale=0.4] at (-11.5, -5.1) {$x_A$ (original)};
    \node[scale=0.4] at (-15.6, -5.1) {$x_A +\left[-\e_1,\e_1\right]$};

    \node[scale=0.7] at (5,14) {Target network};
    \node[scale=0.4] at (3,10.2) {Interval MixUp Image};

    \node[scale=0.6] at (9,4.5) {IBP};
    \node[scale=0.7] at (8.8, 2.3) {.};
    \node[scale=0.5] at (3, 0.9) {linear layer};
    \node[scale=0.5] at (3, -1.7) {ReLU};
    \node[scale=0.5] at (3,-4.5) {linear layer};
    \node[scale=0.5] at (3,-8.8) {logits};

    \node[scale=0.7] at (18.5,14) {Inference};
    \node[scale=0.5] at (19.5,11.8) {Perturbed Image};
    \node[scale=0.5] at (16, 3.5) {$\mathcal{H}( \mathbf{e}_{i} ; \boldsymbol{\Phi} )$};
    \node[scale=0.5] at (18.5,-0.5) {Target};
    \node[scale=0.5] at (18.5,-1.5) {Network};
    \end{tikzpicture}
    \end{center}
    \caption{\our{} uses a hypernetwork to map task-specific embeddings $\boldsymbol{e}_t$ into target models that propagate virtual Interval MixUp hypercubes, enabling robust multi-task learning and certified adversarial resistance. In the Interval MixUp visualization, edge images are perturbed, the center is unperturbed, and the IBP column shows how the input hypercube is transformed across network layers.}
    \label{fig:merf_appro}
\end{figure*}

Despite its importance, this intersection of continual learning and adversarial robustness remains highly underexplored. Existing approaches that attempt to bridge this gap are limited. AIR \cite{zhou2024defense}, for example, uses unsupervised data augmentations to promote robustness during continual learning, but it is only demonstrated on short task sequences and cannot scale to more complex benchmarks. Double Gradient Projection (DGP) \cite{ru2024maintaining} enhances robustness by constraining weight updates; however, it necessitates storing large amounts of per-task gradient information. This makes it impractical for large-scale datasets or modern architectures. Moreover, this storage requirement introduces privacy concerns, especially in domains where task data is sensitive and cannot be retained.

To address these limitations, we introduce a new framework for adversarially robust continual learning. Our method uses a shared hypernetwork to generate task-specific model parameters from compact, trainable embeddings. This architecture enables efficient adaptation to new tasks without requiring replay buffers, gradient logs, or full model snapshots, thus ensuring scalability and privacy. To defend against adversarial inputs, we integrate Interval Bound Propagation (IBP)~\cite{gowal2018effectiveness}, which provides formal robustness guarantees by accounting for worst-case input perturbations. We further introduce Interval MixUp, a novel technique that extends traditional MixUp by interpolating not individual input samples, but uncertainty-aware regions in the input space, enhancing robustness during training. The standard MixUp strategy is unsuitable in our setting, as training on samples with excessively large perturbation radii degrades model performance. To mitigate this, we constrain the hypercube volume near decision boundaries and employ a weighted cross-entropy loss. Empirical results demonstrate that Interval MixUp is essential for achieving high-performing models.

Together, these components form a unified and efficient framework for continual learning in the presence of adversarial threats. By jointly addressing the challenges of continual adaptation and certifiable robustness, our approach takes a significant step toward building safe, trustworthy, and lifelong learning AI systems capable of operating reliably in complex, dynamic environments. An overview of our method is illustrated in Fig.~\ref{fig:merf_appro}.

The key contributions of this work are as follows:
\begin{itemize}
    \item We propose \our{}, a novel architecture for robust continual learning that unifies certified adversarial robustness with strong sequential task performance.

    \item Our method integrates IBP into a hypernetwork architecture and introduces Interval MixUp, a novel training strategy blending virtual interpolation and interval arithmetic to widen decision boundaries and enhance robustness.

    \item \our{} is attack-agnostic and achieves state-of-the-art or highly competitive results on multiple continual learning benchmarks, including Split miniImageNet, with up to 2$\times$ higher adversarial accuracy than prior methods.
\end{itemize}
\section{Related Works}
\label{sec:relatedworks}

Continual learning and adversarial robustness have largely been studied in isolation, with only a few attempts to bridge the two (see Supplementary Material (SM)~\ref{appendix:extended_related_works} for an in-depth overview). Recent work has begun to expose the vulnerabilities of continual learning models to adversarial attacks. For example, \cite{khan2022susceptibility} evaluates several continual learning methods under adversarial scenarios, highlighting the need for joint treatment of robustness and forgetting. AIR~\cite{zhou2024defense} offers one such integrated approach using unsupervised data augmentations for adversarial defense, though its effectiveness is limited to short task sequences.

Gradient-based methods like GEM~\cite{Paz2017gradient} and GPM~\cite{saha2021gradient} primarily aim to prevent forgetting by using episodic memory and subspace projections, respectively, but do not explicitly address adversarial robustness. DGP~\cite{ru2024maintaining}, built as an extension of GPM, further refines these projections to maintain stability across tasks. Interestingly, the authors of DGP observe that while their method improves continual learning performance, the projection strategy itself makes the model more susceptible to adversarial attacks, revealing a critical trade-off between preserving past knowledge and ensuring robustness.

Crucially, none of these methods explore hypernetworks~\cite{ha2016hypernetworks}, despite their proven benefits in continual learning~\cite{henning2021posterior,ksikazek2023hypermask,wang2024comprehensive} for achieving modular task representations.

\section{\our{}: Secure Hypernetworks for Incremental Expansion Learning Defense}

In this section, we introduce \our{}, a model designed for continual adversarial defense across a sequence of tasks. \our{} comprises two primary components: a hypernetwork that generates task-specific parameters and a target network that leverages interval arithmetic to enable certified robustness. We begin by outlining the overall architecture, followed by a formal definition of certified robustness in neural networks. We then demonstrate how \our{} meets these criteria.

\subsection{SHIELD Architecture}\label{sec:arch}

\paragraph{Hypernetwork.} Our model uses trainable embeddings $\boldsymbol{e}_t \in \mathbb{R}^{N}, \mbox{ where } t \in \{ 1, ..., T \}$ represents consecutive tasks indices and $T$ is its total number. These vectors serve as the input to the hypernetwork, which produces the task-dedicated target network weights. More specifically, the hypernetwork $\mathcal{H}$ with weights $\boldsymbol{\Phi}$ generates weights $\boldsymbol{\theta}_t$ of the target network $f$, designed for the $t$--th task, i.e. $\mathcal{H}( \boldsymbol{e}_t ; \boldsymbol{\Phi} ) = \boldsymbol{\theta}_t$. Therefore, the target network is not trained directly, and the meta-architecture generates distinct weights for each continual learning task. The function 
$f_{\boldsymbol{\theta}_t} : X \to Y$ predicts labels $Y$ for data samples $X$ based on weights $\boldsymbol{\theta}_t$. Finally, each continual learning task is represented by a classifier function 
\begin{equation}
f_{\boldsymbol{\theta}_t}\left(\cdot\right) = f(\cdot;\boldsymbol{\theta}_t) = f\big( \cdot ; \mathcal{H}(\boldsymbol{e}_t ; \boldsymbol{\Phi})\big).
\end{equation}
After training, a single meta-model produces distinct task-specific weights, effectively minimizing forgetting.

\paragraph{Target Network.} In \our{}, the target network is designed for interval representation of input points instead of single real values. In practice, we use the IBP approach~\cite{gowal2018effectiveness,morawiecki2019fast} in the target model. Instead of utilizing a single-valued point $x = \left(x_1,\ldots,x_d\right) \in \R^d$ as input to the target model $f_{\boldsymbol{\theta}_t}$, we propagate a $d$-dimensional hypercube:
\begin{equation}
 I_{\e}(x) =  [x_1 - \e, x_1 + \e] \times \ldots \times [x_d - \e, x_d + \e],
\end{equation}

\noindent where $\e \geq 0$. When we consider a target model $f_{\boldsymbol{\theta}_t}$, it is composed of $K$ layers defined by the transformations
\begin{align}
z_k &= h_k(z_{k-1}) = W_k z_{k-1} + b_k, \quad \text{for } k = 2, \ldots , K, \label{eq:affine_transform} \\
z_1 &= x.
\end{align}
In practice, the transformations $h_k$ are connected by non-linear, non-decreasing activation functions such as ReLU or sigmoid, so that each layer consists of an affine mapping followed by a non-linearity. For clarity of exposition, we express each $h_k$ in affine form only in Eq.~\eqref{eq:affine_transform}. In the IBP approach \cite{gowal2018effectiveness}, in each layer, we find the smallest bounding box that encloses the transformed output of the previous layer. To bound the activation $z_k$ of the $k$--th layer in a neural network, we compute an axis-aligned hypercube $[\underline{z}_k, \bar{z}_k]$,
where \( \underline{z}_k \) and \( \bar{z}_k \) represent the element-wise lower and upper bounds, respectively. This hypercube can be written as:
\begin{equation}
I_{\epsilon}(z_k) = [\underline{z}_k, \bar{z}_k] = [\underline{z}_{k,1}, \bar{z}_{k,1}] \times \ldots \times [\underline{z}_{k,d_k}, \bar{z}_{k,d_k}],
\end{equation}
where $d_k$ is the output dimension of the $k$--th layer.

To propagate such intervals efficiently through the layers of a neural network, we use interval arithmetic, leveraging a midpoint-radius representation. This strategy enables efficient and scalable computations. Specifically, by applying Eq.~\eqref{eq:affine_transform} to the input hypercube $[\underline{z}_{k-1}, \bar{z}_{k-1}]$, we define:
\begin{equation}
\begin{split}
    \mu_{k-1} = \frac{\bar{z}_{k-1} + \underline{z}_{k-1}}{2}, & \quad
    r_{k-1} = \frac{\bar{z}_{k-1} - \underline{z}_{k-1}}{2}, \\[0.9ex]
    \mu_k = W_k \mu_{k-1} + b_k, & \quad
    r_k = |W_k| r_{k-1}, \\[0.9ex]
    \underline{z}^\star_k = \mu_k - r_k, & \quad
    \bar{z}^\star_k = \mu_k + r_k,
\end{split}
\end{equation}
where \( |\cdot| \) denotes the element-wise absolute value operator. The resulting hypercube \( [\underline{z}^\star_k, \bar{z}^\star_k] \) serves as the input to the next activation function. For a monotonic non-decreasing activation function \( g(\cdot) \), the final output bounds after applying the activation are given by:
\begin{equation}
\begin{split}
    \underline{z}_k = g(\underline{z}^\star_k), \qquad
    \bar{z}_k = g(\bar{z}^\star_k).
\end{split}
\end{equation}
This procedure, known as IBP, enables tractable, layer-wise propagation of bounds through the entire network.

To ensure robust classification, we adopt a worst-case loss function during training (see Eq.~\eqref{worstcase}) that accounts for the entire interval bound $[\underline{z}_K, \overline{z}_K]$ of the final logits. This encourages the model to make consistent predictions for all inputs within a certified neighborhood of the original sample.

The goal is to guarantee that the model correctly classifies not only the input sample $x$, but also all perturbed samples $\hat{x}$ within the $\e$-bounded $\ell_\infty$ hypercube, i.e., for all $\hat{x} \in [x - \e, x + \e]$. This ensures that the predicted class remains unchanged within a specified neighborhood around the input, providing certified robustness against adversarial perturbations.

Ultimately, the model comprises a learnable embedding and a hypernetwork designed to create the IBP-based target model for a given task. Therefore, \our{} has improved robustness against adversarial attacks. 

\subsection{\our{} Loss Function}

One of the core components of our model is the loss function, which simultaneously mitigates catastrophic forgetting and enhances robustness against adversarial attacks. We adopt a worst-case loss computed over the interval bounds $[\underline{z}_K, \overline{z}_K]$ of the final logits. This approach ensures that the entire output bounding box is classified correctly, meaning that all permissible input perturbations within the specified interval do not alter the predicted class.

Concretely, the logits corresponding to the true class $y_{\text{true}}$ and the predicted class $\hat{y}$ are considered as its lower bound, while the logits of all other classes are set to their respective upper bounds:
\begin{equation}
\hat{z}_{K,\hat{y}} = \left\{
\begin{array}{ll}
     \bar z _K, & \text{for } \hat{y} \neq y_{\text{true}},\\[0.8ex]
     \barbelow z_K, & \text{otherwise}.
\end{array}
\right.\label{worstcase}
\end{equation}

This selection of bounds yields the most pessimistic, interval-based prediction vector, effectively capturing the worst-case scenario within the certified neighborhood. We then apply the softmax function and the cross-entropy loss to $\hat{z}_{K,\hat{y}}$.

As shown in~\cite{gowal2018effectiveness}, computing these interval bounds requires only two forward passes through the network, making the method computationally practical. To address the challenge of overly loose bounds in complex networks, the authors of~\cite{gowal2018effectiveness} proposed a combined loss that blends the standard cross-entropy loss on the nominal inputs with the interval-based worst-case loss:
\begin{equation}
\mathcal{L}_{\text{IBP}} = \kappa \cdot \mathcal{L}_{\text{CE}}(\sigma\left(\hat{y}\right), y_{\text{true}}) + (1-\kappa) \cdot \mathcal{L}_{\text{CE}}(\sigma\left(\hat{z}_{K, \hat{y}}\right), y_{\text{true}}),
\label{IBP}\end{equation}
where $\mathcal{L}_{\text{CE}}$ denotes the cross-entropy loss, $\sigma\left(\cdot\right)$ is a softmax function, and $\kappa \in [0,1]$ balances the trade-off. For the sake of simplicity, we omit arguments of $\mathcal{L}_{\text{IBP}}$. Notably, when $\e = 0$, the interval bounds collapse to point predictions and $\mathcal{L}_{\text{IBP}}$ reduces to the standard cross-entropy loss. To facilitate stable optimization during training, we apply a scheduled annealing of $\kappa$, detailed in SM~\ref{appendix:sec_training_details}. Additionally, we study the effect of different choices of $\kappa$ on the final performance in SM~\ref{appendix:sec_ablation_study}.

Although ensuring adversarial robustness is possible under worst-case accuracy conditions, it is also essential to avoid catastrophic forgetting in the hypernetwork. Consequently, we incorporate a regularization term responsible for maintaining knowledge from previous continual learning tasks, as in~\cite{von2019continual}. Therefore, we can compare the fixed hypernetwork output (i.e. the target network weights) that was generated before learning the current task $t_c \in \lbrace 2, \ldots , T \rbrace$, denoted by $\boldsymbol{\Phi}^{*}$, with the hypernetwork output after recent suggestions for its weight changes, $\boldsymbol{\Phi} + \Delta \boldsymbol{\Phi}$. Ultimately, in \our{}, the formula for the regularization loss is presented as:
\begin{equation}\label{eq:hyper_loss_reg}
\mathcal{L}_{\text{out}} = \frac{1}{t_c - 1} \cdot  \sum_{t=1}^{t_c-1} \| \mathcal{H}(\mathbf{e}_{t}; \boldsymbol{\Phi}^{*}) - \mathcal{H}(\mathbf{e}_{t}; \boldsymbol{\Phi} + \Delta \boldsymbol{\Phi}) \|^2.
\end{equation}
The final loss function consists of the $\mathcal{L}_{\text{IBP}}$ loss and hypernetwork regularization term $\mathcal{L}_{\text{out}}$:
\begin{equation}\label{eq:final_loss}
     \mathcal{L}_{\text{total}} = \mathcal{L}_{\text{IBP}} + \beta \cdot \mathcal{L}_{\text{out}}
\end{equation}
where $\beta > 0$ is a hyperparameter that influences the stability of the hypernetwork.

\subsection{Robustness Guarantees of SHIELD}

In this subsection, we provide theoretical justification for the certified robustness properties achieved via IBP in \our{}. We first formalize certified robustness for standard feedforward networks, then extend the concept to the continual learning setting, and finally apply it to our proposed method.

\begin{definition}[Certified Robustness]\label{def:ibp_cert_robustness}
Let $y_{\text{true}}$ be the true class of an input sample $x \in \mathbb{R}^d$. A classifier $f_{\boldsymbol{\theta}}$ parametrized by $\boldsymbol{\theta}$ is certifiably robust at $x$ under $\ell_\infty$ perturbations of radius $\e$ if:
\begin{equation}
\underline{z}^{(K)}_{y_{\text{true}}}\left(x; \boldsymbol{\theta}\right) > \max_{j \neq y_{\text{true}}}\overline{z}^{(K)}_{j}\left(x; \boldsymbol{\theta} \right),
\end{equation}
\noindent where $\underline{z}^{(K)}$, $\overline{z}^{(K)}$ are lower and upper bounds, respectively, on the output logits for $x$ at the final layer $K$, computed using IBP.
\end{definition}

\begin{definition}[Certified Robustness in Continual Learning]\label{def:ibp_cl_cert_robustness}
Consider a continual learning setup with a sequence of $T$ tasks, $\{\mathcal{D}_t\}_{t=1}^T$, where each task dataset $\mathcal{D}_t = \{(x_i^{(t)}, y_i^{(t)})\}_{i=1}^{N_t}$. Let $\boldsymbol{\theta}_t$ denote the model parameters after learning task $t$. The model is said to be certifiably robust up to time $t$ if, for all samples $(x, y_{\text{true}}) \in \bigcup_{s=1}^{t} \mathcal{D}_s$, the following condition holds:
\begin{equation}
\underline{z}^{(K)}_{y_{\text{true}}}(x; \boldsymbol{\theta}_t) > \max_{j \neq y_{\text{true}}} \overline{z}^{(K)}_j(x; \boldsymbol{\theta}_t).
\label{eq:robustness}\end{equation}
This condition guarantees that the model’s predictions remain robust to bounded adversarial perturbations across all tasks encountered so far.
\end{definition}

To understand how robustness can be preserved across tasks, we analyze the effect of parameter updates during continual learning. When the model begins learning task $(t+1)$, the parameters of the hypernetwork are updated, which in turn alters the classifier weights it generates. We denote by $\boldsymbol{\theta}_{s,t}$ the classifier weights produced by the hypernetwork to solve the $s$-th task after learning task $t$. Consequently, these weights are modified from $\boldsymbol{\theta}_{s,t}$ to $\boldsymbol{\theta}_{s,t} + \boldsymbol{h}$, where $\boldsymbol{h}$ denotes the induced update. To maintain the robustness guarantees established for any previous task $s \le t$, this update must not violate the robustness condition given in Eq.~\eqref{eq:robustness}.

For notational simplicity, when we write $\boldsymbol{\theta}_t$ with a single subscript, we refer to the classifier weights generated to solve the $t$-th task, without specifying the stage of continual training at which they were produced.

We characterize how much the classifier outputs may change under the update $\boldsymbol{h}$ while still preserving the previously certified robustness. This leads to the following sufficient condition for robustness preservation.

\begin{theorem}[Certified Robustness Preservation]\label{thm:robustness_preservation}
Suppose that learning task $(t+1)$ updates the classifier parameters by $\boldsymbol{h}$, yielding $\boldsymbol{\theta}_{s,t} + \boldsymbol{h}$. For any previous task $s \in \{1, \ldots, t\}$, let the certified margin at time $t$ for a sample $(x, y_{\text{true}}) \in \mathcal{D}_s$ be defined as
$
M(x,y_{\text{true}}; \boldsymbol{\theta}_{s,t}) = \underline{z}^{(K)}_{y_{\text{true}}}\left(x; \boldsymbol{\theta}_{s,t}\right) - \max_{j \neq y_{\text{true}}} \overline{z}^{(K)}_{j}\left(x; \boldsymbol{\theta}_{s,t} \right).$

Let $\Delta_{\text{max}}^{(s,t)}(x; \boldsymbol{h})$ denote the maximum change in the classifier’s output logits on $(x, y_{\text{true}})$ due to the parameter update $\boldsymbol{h}$: $\Delta_{\text{max}}^{(s,t)}(x; \boldsymbol{h})
= \| f(x; \boldsymbol{\theta}_{s,t} + \boldsymbol{h}) - f(x; \boldsymbol{\theta}_{s,t}) \|_{\infty}.$ Then, a continual learner preserves certified robustness for $(x, y_{\text{true}}) \in \mathcal{D}_s$ if the parameter update $\boldsymbol{h}$ satisfies
\begin{equation}\label{eq:robustness_assumption}
\Delta_{\text{max}}^{(s,t)}(x; \boldsymbol{h})
\leq \tfrac{1}{2} M(x, y_{\text{true}}; \boldsymbol{\theta}_{s,t}).
\end{equation}
\end{theorem}
Proof is provided in SM~\ref{appendix:sec_theoretical_analysis}.

\our{} unifies IBP-based robust training with a hypernetwork-driven continual learning framework to maintain certified robustness across sequential tasks. Specifically, the hypernetwork generates task-specific parameters $\boldsymbol{\theta}_t = \mathcal{H}(\boldsymbol{e}_t; \boldsymbol{\Phi})$, while IBP training (Eq.~\eqref{IBP}) enforces that each task model remains certifiably robust within its interval bounds. To mitigate catastrophic forgetting, the regularization term in Eq.~\eqref{eq:hyper_loss_reg} enforces consistency with parameters produced for prior tasks, stabilizing learned representations and preserving previously acquired knowledge. Importantly, as stated in Theorem~\ref{thm:robustness_preservation}, when this regularization is sufficiently strong, it ensures not only the retention of past task performance but also the preservation of their certified robustness guarantees. We additionally provide a theoretical analysis of \our{} in SM~\ref{appendix:sec_theoretical_analysis}, formally establishing its robustness-preserving continual learning capabilities.
\section{Enhancing Certified Robustness of \our{} via Interval MixUp}
In this section, we revisit the MixUp technique~\cite{zhang2018mixupempiricalriskminimization} and adapt it to align with interval arithmetic and continual learning. We then demonstrate that our interval-based MixUp approach enhances the robustness of our method by encouraging the decision boundary to move as far as possible from intervals to be correctly classified, without sacrificing accuracy.

\paragraph{MixUp.}
MixUp~\cite{zhang2018mixupempiricalriskminimization} improves model generalization by training on convex combinations of input samples and their corresponding labels. Instead of optimizing the model on individual examples, MixUp introduces synthetic points that lie along linear paths between pairs of training samples. This encourages the model to behave smoothly between classes and to avoid overly confident or abrupt predictions in regions where data support is lacking.

MixUp introduces an inductive bias that promotes linear behavior between training examples. It smooths the model's decision surface, reduces memorization of noisy labels, and improves generalization. Enforcing consistency across interpolated inputs and outputs also enhances robustness to ambiguous or uncertain regions in the input space. Additional technical details on MixUp are provided in SM~\ref{appendix:sec_mixup}. 

\paragraph{Interval MixUp.}

To enable the benefits of MixUp in the context of certified training, we introduce an extension called Interval MixUp, which operates directly on interval-bounded inputs as defined in Interval Bound Propagation (IBP). Unlike standard MixUp that interpolates between individual input points, Interval MixUp interpolates entire $\ell_\infty$-bounded perturbation sets, represented as hypercubes, resulting in virtual inputs with smooth certified regions and continuous transitions across class boundaries.

Importantly, our method deviates from the standard MixUp formulation by avoiding label interpolation. Instead, it performs interpolation at the loss level by computing the cross-entropy loss on virtual inputs and weighting the individual loss terms using the MixUp coefficient. This leads to a principled formulation that remains compatible with the requirements of certified training.


First, we construct a virtual sample by linearly interpolating between the midpoints of two original $\ell_\infty$-bounded hypercubes. This yields a new hypercube centered at the interpolated point, which remains a valid input region and can be propagated through the network using standard IBP-based methods. To complete the definition of this virtual region, we must also specify its radius.

Optimizing the IBP loss under large perturbation radii $\e$ is difficult due to unstable gradients, overly loose interval bounds, and excessive over-approximation of the certified region. These challenges often result in degraded training dynamics and suboptimal convergence. However, applying large perturbations to virtual points that lie far from the original data manifold is unnecessary and can even be harmful. To address this, we propose a simple yet effective strategy: define the radius of the new hypercube as a function of its distance from the original data. Specifically, we reduce the perturbation radius as the interpolated point moves farther away from its endpoints. Let $\lambda \in [0, 1]$ denote the MixUp mixing coefficient. The adjusted perturbation radius is then defined as:
\begin{equation}\label{eq:eps_scaled}
    \e' = |2\lambda - 1| \cdot \e,
\end{equation}
which smoothly scales the certified region depending on the location of the virtual sample between its source points.

This formulation yields the maximum radius when $\lambda$ is close to 0 or 1, i.e., when the virtual sample is close to one of the original points, and reduces it to zero when $\lambda = 0.5$, i.e., when the sample is equidistant and lies furthest from either original data point. This leads to a smoothed decision boundary and more stable training while maintaining robustness near the data manifold. In SM~\ref{appendix:sec_ablation_study}, we explore the influence of different transformations in Equation \eqref{eq:eps_scaled} (such as linear, quadratic, logarithmic, and cosine ones) on the training process and defense against adversarial attacks.

Furthermore, both the forward and backward passes in Interval MixUp are performed solely on virtual, interval-bounded examples obtained by interpolating pairs of inputs. As a result, the propagated hypercubes (i.e., interval bounds around the inputs) have smaller radii than those centered at the original data points. This has a crucial side effect: it mitigates the wrapping effect~\cite{neumaier1993wrapping}, a well-known limitation of interval arithmetic. The wrapping effect arises when high-dimensional intervals are passed through non-linear layers of the neural network, causing them to grow excessively and become increasingly misaligned with the true shape of the reachable set. This over-approximation results in looser, overly conservative bounds that may significantly underestimate the model’s actual robustness.

\begin{figure}[!ht]
    \centering
        \includegraphics[width=0.48\linewidth]{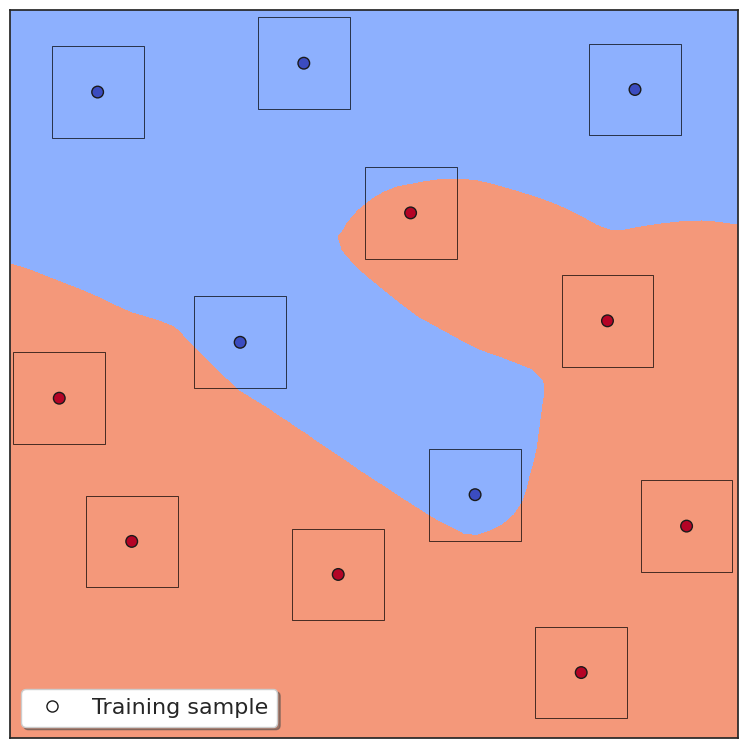}
        \includegraphics[width=0.48\linewidth]{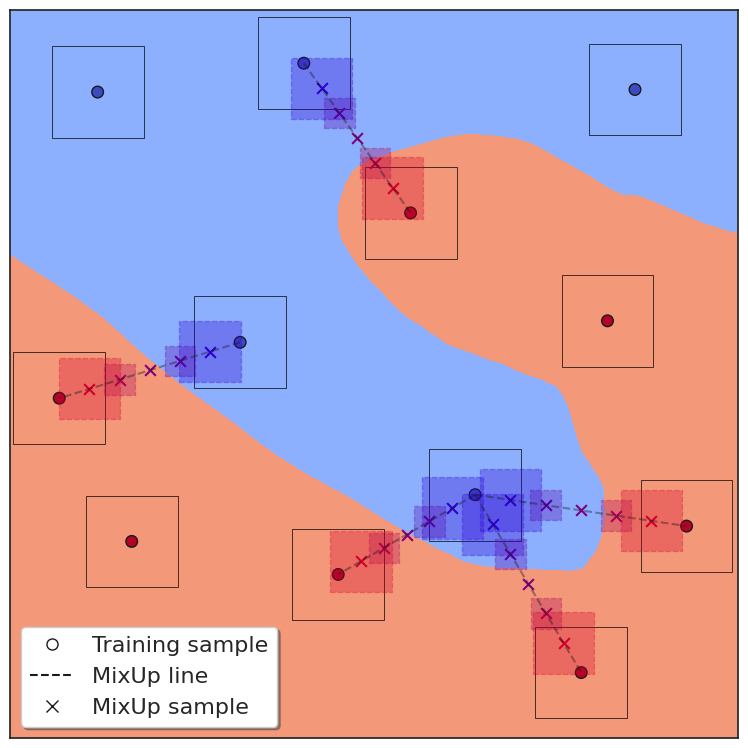}
    \caption{(left side) Without Interval MixUp: Training leads to sharper boundaries and poor robustness across class transitions. (right side) With Interval MixUp: Interpolated samples ($\times$) with scaled certified regions (boxes) encourage smooth transitions and robust boundaries.
    }
    \label{fig:interval_mixup}
\end{figure}

To build intuition for how Interval MixUp operates in practice, we construct a 2D toy experiment, visualized in Fig.~\ref{fig:interval_mixup}. We generate synthetic training data consisting of a few points from two distinct classes, randomly placed within the unit square $[0,1]^2$. Each point is associated with a certified region, shown as a square representing an $\ell_\infty$-bounded hypercube.

We then identify several pairs of nearby points belonging to opposite classes. For each such pair, we apply Interval MixUp by linearly interpolating the bounds of their certified regions using a mixing coefficient $\lambda \in [0,1]$. This produces a set of virtual samples along the line connecting each pair. These samples are visualized as $\times$ markers colored by their interpolation ratio, ranging from red (class 0) to blue (class 1), with intermediate hues indicating mixed contributions. Each virtual point is surrounded by a semi-transparent certified box, whose size is adjusted using Eq.~\eqref{eq:eps_scaled}. We train \our{} model on this setup using only the Interval MixUp samples. The resulting decision surface and certified regions, shown in Fig.~\ref{fig:interval_mixup}, illustrate how Interval MixUp helps populate the space between classes, guiding the decision boundary away from both the real data. 

\begin{figure}[t]
    \centering
    \includegraphics[trim={0,2cm, 0,1cm, 1,5cm, 0,2cm},clip,width=0.49\linewidth]{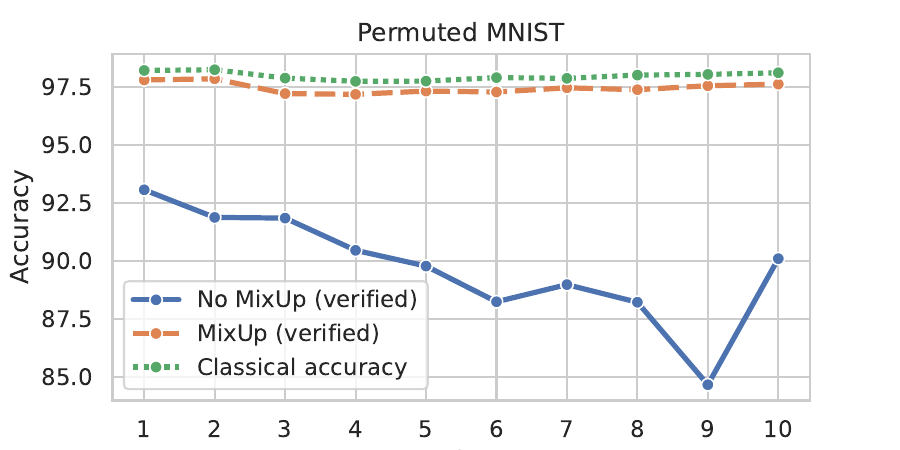}%
    \includegraphics[trim={0,2cm, 0,1cm, 1,5cm, 0,2cm},clip,width=0.49\linewidth]{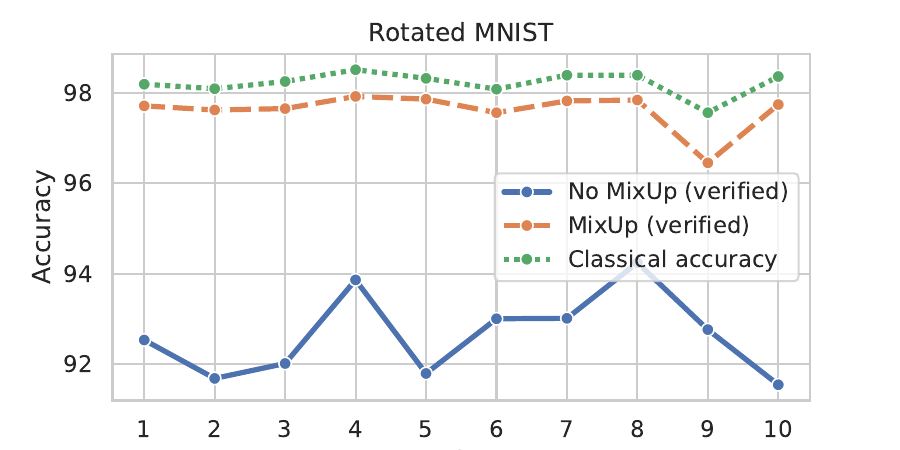}
    \includegraphics[trim={0,2cm, 0,1cm, 1,5cm, 0,2cm},clip,width=0.49\linewidth]{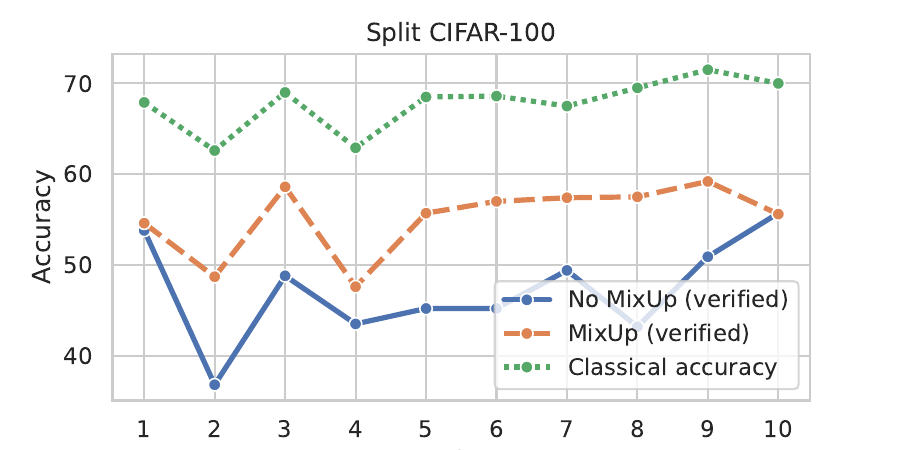}%
    \includegraphics[trim={0,2cm, 0,1cm, 1,5cm, 0,2cm},clip,width=0.49\linewidth]{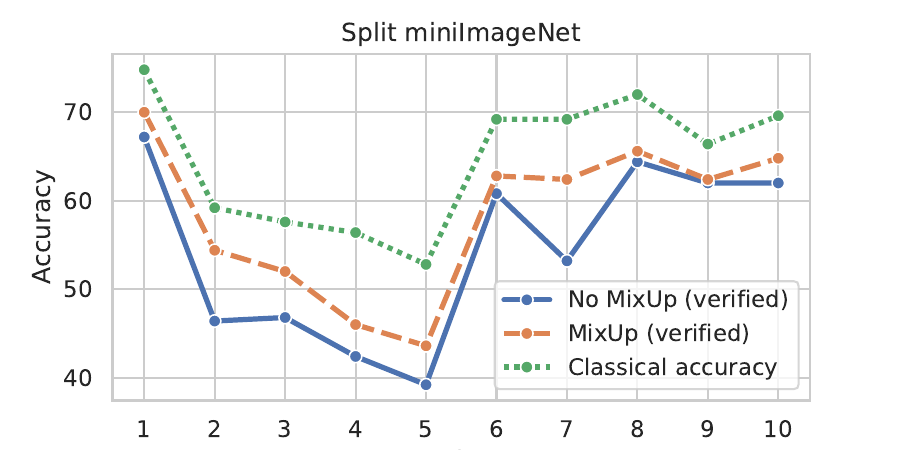}

    \caption{
Comparison of verified accuracy and classical accuracy across four continual learning benchmarks:  Permuted~MNIST, Rotated~MNIST, Split~CIFAR-100, and  Split~miniImageNet. For each task, we report the AA metric after sequentially learning all preceding tasks.
}\label{fig:ver_acc}
\end{figure}

By interpolating smaller virtual hypercubes, Interval MixUp reduces the growth of these intervals during propagation, thereby maintaining tighter and more informative bounds. This is illustrated in Fig.~\ref{fig:ver_acc}, which reports verified accuracy~\cite{zhang2018efficientneuralnetworkrobustness}. Verified accuracy measures whether a model is not only correct on a given input but also robust to all perturbations within a specified $\e$-ball around it, unlike classical accuracy, which only evaluates predictions on clean, unperturbed data. Verified accuracy thus reflects a model’s certified robustness by leveraging the interval bounds computed during training.

Importantly, models trained with Interval MixUp achieve significantly higher verified accuracy compared to those trained with standard IBP, and their verified accuracy closely matches their classical accuracy (see Figure \ref{fig:ver_acc}). 
As a consequence, these models outperform standard IBP baselines and are more resilient to adversarial attacks evaluated in this paper. All reported results are based on the best-performing models selected using a single random seed. Moreover, we present examples of Interval MixUp samples in SM~\ref{appendix:sec_interval_mixup_samples}.

\paragraph{\our{} Loss Function with Interval MixUp.}

When applying Interval MixUp in \our{}, we cannot directly use Eq.~\eqref{IBP}, as the original input data is replaced by virtual samples generated through the interpolation of input pairs. To accommodate this, we introduce a reformulated loss function designed specifically for these synthetic samples.

Following the standard MixUp framework, we first adapt the cross-entropy loss to account for mixed labels. The MixUp loss is defined as:
\begin{equation}\label{eq:mixup_loss}
\begin{aligned}
    \mathcal{L}_{\text{MixUp}}\left(\tilde{x}, y_a, y_b \right) 
    &= \lambda \, \mathcal{L}_{\text{CE}}\left(\sigma\left(f\left(\tilde{x}\right); \boldsymbol{\theta}_t\right), y_a\right) \\
    &+ \left(1 - \lambda\right) \, \mathcal{L}_{\text{CE}}\left(\sigma\left(f\left(\tilde{x}\right); \boldsymbol{\theta}_t\right), y_b\right)
\end{aligned}
\end{equation}
where $f(\cdot; \boldsymbol{\theta}_t)$ denotes the neural network, $\tilde{x} = \lambda x_a + (1 - \lambda) x_b$ is the interpolated input sample, and $y_a$, $y_b$ are the corresponding one-hot labels for $x_a$ and $x_b$, respectively. To integrate certified robustness into this formulation, we extend the loss to an interval-based version. The resulting Interval MixUp loss is:
\begin{equation}\label{eq:interval_mixup_loss}
\begin{aligned}
    \mathcal{L}_{\text{IMixUp}} &= \kappa \cdot \mathcal{L}_{\text{MixUp}}\left(\tilde{x}, y_a, y_b \right) \\ 
    &+ (1 - \kappa) \cdot \hat{\mathcal{L}}_{\text{MixUp}}\left([\tilde{x} - \e', \tilde{x} + \e'], y_a, y_b \right),
\end{aligned}
\end{equation}
where $\e'$ is the scaled perturbation radius (see Eq.~\eqref{eq:eps_scaled}), and $\kappa \in [0,1]$ balances the standard and certified components of the loss. We omit arguments of $\mathcal{L}_{\text{IMixUp}}$ to simplify the notation. The term $\hat{\mathcal{L}}_{\text{MixUp}}$ is defined as:
\begin{equation}
\begin{aligned}
    \hat{\mathcal{L}}_{\text{MixUp}}& \left([\tilde{x} - \e', \tilde{x} + \e'], y_a, y_b \right) = \lambda \cdot \mathcal{L}_{\text{CE}}\left(\sigma\left(\hat{z}_{K,\hat{y}_a}\right), y_a \right)\\ 
    &+ (1 - \lambda) \cdot \mathcal{L}_{\text{CE}}\left(\sigma\left(\hat{z}_{K,\hat{y}_b}\right), y_b \right),
\end{aligned}
\end{equation}
where $\hat{z}_{K,\hat{y}_a}$ and $\hat{z}_{K,\hat{y}_b}$ denote the components of the target network's interval output associated with classes $\hat{y}_a$ and $\hat{y}_b$, respectively, after $K$ transformations, as specified in Eq.~\eqref{worstcase}. This formulation enables us to retain certified robustness guarantees even for interpolated virtual samples, allowing us to benefit from MixUp's generalization while maintaining interval-based certification.

Finally, to adapt Interval MixUp to the continual learning setting, we incorporate it into the overall objective by combining it with the hypernetwork regularization term from Eq.~\eqref{eq:hyper_loss_reg}. The complete loss function becomes:
\begin{equation}
    \mathcal{L}_{\text{total}} = \mathcal{L}_{\text{IMixUp}} + \beta \cdot \mathcal{L}_{\text{out}},
\end{equation}
where $\beta$ is a hyperparameter controlling the strength of the hypernetwork regularization, and $\mathcal{L}_{\text{out}}$ is defined as in Eq.~\eqref{eq:hyper_loss_reg}. This final formulation ensures that our model can learn robust, certified representations in the context of continual learning. The training algorithm of our method is provided in SM~\ref{appendix:sec_train_algo}.

\section{Experiments}
This section presents the experimental setup and results for \our{}. We first describe the datasets and the architectures of both the target network and the hypernetwork. All experiments are conducted in the Task-Incremental Learning setting, where the task identity is provided to the model during both training and evaluation. In addition, our method naturally extends to the Class-Incremental Learning (CIL) scenario while retaining a significant degree of adversarial robustness. SM~\ref{appendix:sec_cil} shows that \our{} consistently outperforms a strong baseline under multiple attack settings in the CIL scenario.

Since AIR~\cite{zhou2024defense} is not attack-agnostic and therefore operates under a different set of assumptions than \our{}, we include a separate comparison in SM~\ref{appendix:sec_air_comparison}. Notably, \our{} outperforms AIR despite being task-agnostic, whereas AIR is not.

\paragraph{Benchmarks.}
To assess the effectiveness of our approach, we perform experiments on four standard benchmarks: Permuted MNIST, Rotated MNIST, Split CIFAR-100, and Split miniImageNet. The structure of tasks and class divisions follows the protocol established in \cite{ru2024maintaining}. Full dataset and task details are provided in SM~\ref{appendix:sec_benchmark_datasets}. 

\paragraph{Architectures.}
For Permuted MNIST, Rotated MNIST, and Split miniImageNet, we adopt the same target network architectures as in~\cite{ru2024maintaining}, with key modifications: we introduce interval-aware variants of normalization and pooling layers, include biases, and discard the use of separate convolutional layers per task. For Split CIFAR-100, we use a streamlined interval-based version of AlexNet. Further architectural details are provided in SM~\ref{appendix:sec_training_details}.

\paragraph{Metrics.}  
We evaluate performance using two standard continual learning metrics: Average Accuracy (AA) and Backward Transfer (BWT). AA is computed as $\frac{1}{T} \sum_{t=1}^{T} R_{T,t}$, where $R_{t,s}$ is the test accuracy on task $s$ after training $t$ tasks, where $s\leq t$. BWT is defined as $\frac{1}{T - 1} \sum_{t=1}^{T - 1} (R_{T,t} - R_{t,t})$, measuring the influence of learning new tasks on performance over previously learned ones. To assess continual learning performance, we evaluate accuracy on the test sets of all prior tasks.

\paragraph{Results.}
Tab.~\ref{tab:results} reports the AA and BWT obtained by \our{} and several baselines after sequential learning over all tasks in the Rotated MNIST, Split CIFAR-100, and Split miniImageNet.

\begin{table}[!ht]
\centering
\caption{AA performance after all tasks on the Rotated MNIST, Split CIFAR-100, and Split miniImageNet. Results for \our{} and \our{} with Interval MixUp (\our{}\textsubscript{IM}) are averaged over 2 seeds for AutoAttack and 5 seeds for the other evaluations, except for Split miniImageNet (3 seeds, 2 for AutoAttack). Baseline results are from \cite{ru2024maintaining}. Extended results with standard deviations are in SM~\ref{appendix:sec_extended_results}.}
\label{tab:results}
{\small
\begin{tabular}{@{}l@{}c@{}c@{}c@{}c@{}c@{}c@{}c@{}c@{}}
\hline
\multicolumn{1}{c}{Method} & \multicolumn{1}{c}{AutoAttack} & \multicolumn{1}{c}{PGD} & \multicolumn{1}{c}{FGSM} & \multicolumn{2}{c}{Original samples} \\ \hline 

\multicolumn{1}{c}{Metric} & AA(\%) & AA(\%) & AA(\%) & AA(\%) & BWT \\ 
\hline


\multicolumn{6}{c}{Rotated MNIST} \\ 
\hline
SGD & 14.1 & 9.9  & 20.4 & 32.3 & -0.71 \\
SI & 13.9  & 15.3  & 20.1 & 33.0 & -0.72 \\
A-GEM & 14.1  & 21.6  & 24.8 & 45.4 & -0.57 \\
EWC & 45.1  & 49.5  & 46.5  & 80.7 & -0.18 \\
GEM & 11.9  & 76.5  & 74.4  & 96.7 & -0.01 \\
OGD & 19.7  & 23.8  & 23.8  & 48.0 & -0.55 \\
GPM & 68.8  & 71.5  & 65.9  & 97.1 & -0.01 \\
DGP & 81.6 & 82.6 & 78.6 & 98.1 & \textbf{-0.00} \\ \hline
\our{} & \textbf{85.64} & 92.94 &  \textbf{83.82} & 95.62 & -0.03 \\
\our{}\textsubscript{IM} & 82.91 & \textbf{97.88} & 83.03 & \textbf{98.32} & -0.08 \\
\hline
\multicolumn{6}{c}{Split CIFAR-100} \\
\hline
SGD & 10.3 & 12.8 & 19.4 & 46.5 & -0.49 \\
SI & 13.0 & 15.2 & 19.8 & 45.4 & -0.48 \\
A-GEM & 12.6 & 12.9 & 20.7 & 40.6 & -0.48 \\
EWC & 12.6 & 23.2 & 30.5 & 56.8 & -0.35 \\
GEM & 21.2 & 19.4 & 47.7 & 60.6 & -0.13 \\
OGD & 11.8 & 14.1 & 18.9 & 44.2 & -0.50 \\
GPM & 34.4 & 36.6 & \textbf{53.7} & 58.2 & \textbf{-0.10} \\
DGP & 36.6 & 39.2 & 48.0 & 67.2 & -0.13 \\ \hline
\our{} & 60.91 & 59.77 & 45.37 & 64.24 & -0.34 \\ 
\our{}\textsubscript{IM} & \textbf{63.08} & \textbf{62.39} & 46.48 & \textbf{67.45} & -0.41 \\ \hline
\multicolumn{6}{c}{Split miniImageNet} \\ 
\hline
SGD   & 20.5 & 22.0 & 23.5 & 30.8 & -0.24 \\
A-GEM & 19.0 & 19.8 & 21.2 & 29.2 & -0.28 \\
EWC   & 21.3 & 22.7 & 24.3 & 29.9 & -0.25 \\
SI    & 20.4 & 21.3 & 22.7 & 28.1 & -0.27 \\
GEM   & 22.3 & 23.8 & 25.4 & 31.8 & -0.20 \\
OGD   & 17.9 & 18.8 & 20.7 & 29.6 & -0.29 \\
GPM   & 26.3 & 27.1 & 28.8 & 36.8 & -0.12 \\
DGP   & 32.1 & 33.8 & 35.5 & 44.8 & \textbf{-0.05} \\\hline
\our{} & 56.22 & 56.8 & 53.08 & 59.52 & -0.16 \\
\our{}\textsubscript{IM} & \textbf{57.9} & \textbf{58.47} & \textbf{54.1} & \textbf{62.67} & -0.18 \\
\hline
\end{tabular}
}
\end{table}

On Rotated MNIST, \our{} outperforms all baselines across all metrics, including AutoAttack, PGD, FGSM, and clean accuracy. \our{}\textsubscript{IM} enhances PGD and clean performance further, demonstrating robustness to rotation-based shifts. We also report results on Permuted MNIST in SM (Sec.~\ref{appendix:sec_ablation_study}), where \our{} achieves state-of-the-art performance.

On the more complex Split CIFAR-100 benchmark, \our{} again achieves the top AA under AutoAttack and PGD. While its original sample AA and BWT are slightly below GPM and DGP, they improve with the Interval MixUp variant, confirming scalability to harder tasks while maintaining strong adversarial performance.

Across benchmarks, \our{} demonstrates a strong balance between adversarial robustness and generalization. This balance is non-trivial due to the well-known trade-off: larger training $\e$ improves robustness under AA and PGD, but can degrade performance on original samples due to increased regularization. \our{}\textsubscript{IM} effectively mitigates this issue, consistently improving both adversarial and original sample AA. The relatively low FGSM results are explained by a mismatch between test-time $\e_{\text{attack}}$ and the training $\e$.

As shown in Tab.~\ref{tab:results}, on Split miniImageNet, \our{} significantly outperforms all baselines across all metrics. While DGP and GPM reach original sample AA of 44.8\% and 36.8\%, respectively, \our{} achieves 59.52\%, and \our{}\textsubscript{IM} reaches 62.67\%. It achieves the best AA on every task and the overall average. Forgetting (measured by BWT) is also well-controlled, with \our{} and \our{}\textsubscript{IM} reaching -0.16 and -0.18, respectively, while achieving far higher AA.

Taken together, the results confirm that \our{} supports robust and scalable continual learning. It maintains and improves performance over time, builds adversarially robust representations, and generalizes well across diverse tasks. Interval MixUp further improves the robustness–accuracy trade-off, making \our{}\textsubscript{IM} a practical and effective solution under adversarial threat models.
\section{Conclusions}
\our{} is the first certifiably robust continual learning method, combining hypernetworks with interval arithmetic to ensure robustness across tasks. Its core component, Interval MixUp, assigns smaller perturbation radii to virtual points farther from real data, tightening bounds and pushing decision boundaries away from certified regions to enhance robustness and accuracy even under large perturbations.

Moreover, \our{} is the first approach to demonstrate substantial certified robustness in the CIL setting, showing that continual adaptation and formal guarantees can be unified, an important step toward robust lifelong learning.
\paragraph{Limitations.}
While \our{} achieves strong certified robustness and adaptability, the use of IBP, though efficient, may yield overly conservative (i.e., wide) output bounds. This can affect both robustness and computational efficiency, particularly in high-dimensional settings.
\section*{Acknowledgments}
The work of P. Spurek was supported by the National Centre of Science (Poland) Grant No. 2023/50/E/ST6/00068. The work of K. Książek was funded from the flagship project entitled ”Artificial Intelligence Computing Center Core Facility” from the DigiWorld Priority Research Area within the Excellence Initiative - Research University program at Jagiellonian University in Krakow.
{
    \small
    \bibliographystyle{ieeenat_fullname}
    \bibliography{main}
}

\clearpage
\setcounter{page}{1}
\maketitlesupplementary
\appendix

\section{Extended related works}\label{appendix:extended_related_works}

This work introduces a novel synthesis of two previously disjoint research areas in deep learning: continual learning and adversarial robustness. Since existing literature rarely addresses their intersection, we organize the related work section around both foundations to highlight the unique contributions of our method.

\paragraph{Adversarial Attack.}
Adversarial attacks on deep neural networks are typically categorized as white-box or black-box, depending on the attacker's access to the target model. In white-box settings, the adversary fully knows the model, including its architecture, parameters, and gradients. Prominent white-box attacks include gradient-based methods such as FGSM \cite{goodfellow2014explaining}, BIM \cite{huang2011adversarial}, MIM \cite{dong2018boosting}, PGD \cite{madry2017towards}, and AutoAttack \cite{de2021continual}, as well as optimization-based approaches like the OnePixel \cite{su2019one}. 

In black-box settings, attackers have limited access to the model and must rely on indirect signals. Some approaches use confidence scores to estimate gradients, such as zoo \cite{chen2017zoo}, while others operate with only the predicted label, as seen in decision-based methods like the Boundary Attack \cite{brendel2017decision}. Another strategy involves crafting adversarial examples on a surrogate model and transferring them to the target model, exploiting the cross-model transferability of adversarial perturbations \cite{dong2018boosting,dong2019evading}.

Simultaneously with adversarial attack techniques, there are many defence algorithms. Early adversarial defenses relied on heuristics such as input transformations \cite{xie2017mitigating}, model ensembles \cite{bagnall2017training}, and denoisers \cite{jin2019ape}, but many were later found to rely on obfuscated gradients \cite{athalye2018obfuscated}. More robust approaches like adversarial training (AT) \cite{goodfellow2014explaining,jia2022adversarial} and defensive distillation \cite{goldblum2020adversarially,zhu2021reliable} have become dominant. Recent work has enhanced AT with learnable strategies \cite{jia2022adversarial}, feature perturbation \cite{mustafa2020deeply}, and optimization techniques \cite{pang2020bag}. However, the above models are typically static and struggle against evolving attack sequences. Test-Time Adaptation Defense \cite{shi2021online,yang2022adaptive} addresses adaptation to new attacks but neglects previous ones. 

Most existing methods aim to improve adversarial robustness, but they do so without providing formal guarantees. Interval Bound Propagation (IBP)~\citep{gowal2018effectiveness,mirman2018differentiable} offers a principled alternative by using interval arithmetic to train models that are provably robust against adversarial attacks. These methods enforce that the model classifies all points within a specified perturbation region (typically a ball in $\ell_{\infty}$) correctly, by optimizing a worst-case cross-entropy loss.


    
    

\paragraph{Continual Learning.}

Continual learning methods are often grouped into architectural, rehearsal, and regularization-based strategies \cite{hsu2018re}. Architectural approaches like Progressive Neural Networks \cite{rusu2016progressive} and CopyWeights with Re-init \cite{lomonaco2017core50} prevent forgetting by expanding or modifying network structures. Further developments reuse a fixed architecture with iterative pruning, such as PackNet \cite{mallya2018packnet} or Supermasks \cite{wortsman2020supermasks}. Rehearsal-based methods, such as the buffer strategy in \cite{hayes2019memory}, retain selected past examples, while pseudo-rehearsal approaches train generative models to replay prior data \cite{mazur2022target}. Regularization-based techniques like EWC \cite{kirkpatrick2017overcoming}, SI \cite{zenke2017continual}, LwF \cite{li2017learning}, and MAS \cite{aljundi2018memory} constrain parameter updates based on past task importance, often using the Fisher Information Matrix or similar metrics. 

Hypernetworks, introduced in \citep{ha2016hypernetworks}, are defined as neural models that generate weights for a separate target network that solves a specific task. 
The authors aim to reduce the number of trainable parameters by designing a hypernetwork that often has fewer parameters than the target network. In continual learning, hypernetworks can directly generate individual weights for subsequent continual learning tasks, like in HNET~\cite{von2019continual}. Then, HNET was extended with a probabilistic approach called posterior meta-replay~\cite{henning2021posterior}. In HyperMask \cite{ksikazek2023hypermask}, the authors used the lottery ticket hypothesis to produce task-specific masks for the trainable or fixed target model. In \cite{krukowski2024hyperinterval}, the authors employ interval arithmetic on weights to regularize the target model. These methods are positioned between those based on architecture and those focused on regularization.

The above concepts have many different modifications and upgrades \cite{wang2024comprehensive}. But the adversarial robustness of continual learning problems is unexplored in the literature. 
AIR~\cite{zhou2024defense} introduced the first framework for continual adversarial defense, addressing robustness against evolving attack sequences. The method leverages unsupervised data augmentation to enhance general robustness in a task-agnostic manner. While innovative, AIR is limited in scalability, having been evaluated on only two to three defense tasks, raising concerns about its effectiveness in longer, continual settings.

Alternatively, the study on Double Gradient Projection (DGP) \cite{ru2024maintaining} highlights that this technique is notably more resilient to adversarial attacks compared to traditional continual learning approaches such as GEM~\cite{Paz2017gradient} and GPM~\cite{saha2021gradient}. DGP enhances adversarial robustness by projecting gradients orthogonally concerning a critical subspace before updating weights, thus maintaining the prior gradient smoothness from earlier samples. Furthermore, \cite{khan2022susceptibility} examines the resistance of current models against adversarial assaults. These methodologies enhance the robustness of continual learning, yet they have several significant limitations. Firstly, such approaches are restricted to several subsequent tasks and do not give strict guarantees. 
\section{MixUp}\label{appendix:sec_mixup}
\begin{definition}[MixUp]\label{appendix:def_mixup}
Given two training examples $(x_i, y_i)$ and $(x_j, y_j)$ drawn independently from the data distribution, MixUp constructs a new virtual training example $(\tilde{x}, \tilde{y})$ as:
\begin{align}
    \tilde{x} &= \lambda x_i + (1 - \lambda) x_j, \\
    \tilde{y} &= \lambda y_i + (1 - \lambda) y_j,
\end{align}
where $\lambda \in [0,1]$ is sampled from the Beta distribution, $\lambda \sim \text{Beta}(\alpha, \alpha)$, with a fixed hyperparameter $\alpha > 0$.
\end{definition}

Here, $y_i, y_j \in \mathbb{R}^M$ are typically one-hot encoded class labels, so the mixed label $\tilde{y}$ becomes a soft target, a convex combination of label vectors. To train a model with such soft labels, the loss function must support probabilistic targets. A common choice is the soft-label cross-entropy loss:
\begin{equation}
    \mathcal{L}(\tilde{y}, \hat{y}) = - \sum_{k=1}^M \tilde{y}_k \log \hat{y}_k,
\end{equation}
where $\hat{y}$ denotes the model’s predicted probability distribution over classes. This formulation encourages linear behavior between training examples, often improving generalization and calibration.

\section{Interval Neural Network Layers}  
The implementation of linear and convolutional layers in the interval version is straightforward and follows the approach described in the main part. However, implementing other interval layers requires more care and is not as trivial.

\paragraph{Interval Batch Normalization.}  
Batch normalization in the interval version is more involved. Instead of simply normalizing the lower and upper bounds separately, we first concatenate the lower and upper bounds into a single tensor. We then compute the batch statistics, mean, and variance, over this combined tensor, capturing the distribution of the entire interval. Using these statistics, we apply the standard batch normalization transformation.

More concretely, given pre-activation interval bounds $[\underline{x}, \overline{x}]$, we form the concatenated tensor:
\begin{equation}
X_\text{concat} = [\underline{x}, \overline{x}],
\end{equation}
and compute the expected mean $\mu$ and variance $\sigma^2$ over $X_\text{concat}$. We then normalize and scale the interval bounds as follows:
\begin{equation}
\left[\frac{\underline{x} - \mu}{\sqrt{\sigma^2 + \e}}, \frac{\overline{x} - \mu}{\sqrt{\sigma^2 + \e}}\right],
\end{equation}
followed by the affine transformation with learned parameters $\gamma$ (scale) and $\beta$ (shift):
\begin{equation}
\gamma \cdot \left[\frac{\underline{x} - \mu}{\sqrt{\sigma^2 + \e}}, \frac{\overline{x} - \mu}{\sqrt{\sigma^2 + \e}}\right] + \beta.
\end{equation}
We add a small positive constant $\epsilon$ to the denominator to ensure numerical stability. Special attention is required to handle the sign of $\gamma$ correctly to maintain valid lower and upper bounds. If the sign of $\gamma$ is negative, the lower and upper bounds must be swapped after scaling to preserve correct interval ordering.

\paragraph{Interval Pooling Layers.}  
Pooling operations can be extended to the interval version by applying them separately to the lower and upper bounds of each input element. 

For max pooling, given a set of input intervals $\{[\underline{x}_i, \overline{x}_i]\}_{i=1}^n$, the output interval is computed as:
\begin{align}
    \mathrm{MaxPool}_{\inf} &= \max_{1 \leq i \leq n} \underline{x}_i, \\
    \mathrm{MaxPool}_{\sup} &= \max_{1 \leq i \leq n} \overline{x}_i,
\end{align}
and the resulting output interval is $[\mathrm{MaxPool}_{\inf}, \mathrm{MaxPool}_{\sup}]$.

For average pooling, the interval is computed by averaging the bounds independently:
\begin{align}
    \mathrm{AvgPool}_{\inf} &= \frac{1}{n} \sum_{i=1}^n \underline{x}_i, \\
    \mathrm{AvgPool}_{\sup} &= \frac{1}{n} \sum_{i=1}^n \overline{x}_i,
\end{align}
so that the final interval is $[\mathrm{AvgPool}_{\inf}, \mathrm{AvgPool}_{\sup}]$.

Average pooling often produces tighter and more stable bounds, and is therefore preferred in interval-based analysis such as IBP.

\section{Benchmark Datasets}\label{appendix:sec_benchmark_datasets}
Permuted MNIST introduces a different fixed pixel permutation for each task, applied to the original MNIST images. Rotated MNIST, on the other hand, creates tasks by rotating the digits by varying angles. These two datasets are widely used in the continual learning literature as synthetic benchmarks derived from MNIST \cite{goodfellow2014explaining, liu2022continuallearningrecursivegradient}. Split CIFAR-100 \cite{zenke2017continual} is constructed by randomly partitioning the 100 classes of the CIFAR-100 dataset into 10 mutually exclusive groups, with each group forming a separate classification task of 10 classes. Similarly, Split miniImageNet is generated by selecting a subset of the original ImageNet dataset \cite{chaudhry2020continuallearninglowrankorthogonal} and dividing it into 20 distinct groups, each containing 5 unique classes. For both Split CIFAR-100 and Split miniImageNet, each class appears in only one group, ensuring no class overlap between tasks. This design allows each subset to be treated as an independent classification task, resulting in a sequence of tasks for continual learning. To match the setup of Split miniImageNet presented in \cite{ru2024maintaining}, we split Split miniImageNet as described; however, we use only the first ten tasks. Therefore, we have 10 tasks, each with 5 classes.

\section{Discussion on Learnable Number of Parameters Used in \our{}}\label{appendix:sec_learanble_params}
Since we employ hypernetworks to generate the weights of the target network, the total number of learnable parameters is typically higher than that of baseline architectures. However, we demonstrate that it is possible to significantly reduce this parameter count, bringing it in line with standard models, while still achieving state-of-the-art performance. Importantly, the number of parameters in the target network remains comparable to, or even smaller than, those of the baseline models; the additional overhead stems solely from the hypernetwork. The Interval MixUp technique is also not used in this experiment.

For the Permuted MNIST dataset, we reduce the number of learnable parameters to match the budget used in \cite{ru2024maintaining} by employing a hypernetwork composed of a two-layer MLP with hidden dimensions 100 and 50, and an embedding size of 96. The target network is a two-layer MLP with 50 neurons per layer. For the Rotated MNIST dataset, we adopt the same architecture but reduce the embedding size to 24 to further minimize the parameter count. We use $\e = \frac{25}{255}$ for both Permuted MNIST and Rotated MNIST during training.

For the Split CIFAR-100 dataset, again, we align with the parameter budget from \cite{ru2024maintaining} by using an embedding size of 512 and a hypernetwork consisting of a two-layer MLP with 200 and 50 hidden units, respectively. Additionally, we simplify the target AlexNet architecture by removing two fully connected layers, applying a $4 \times 4$ average pooling operation after the final convolutional layer in place of the max pooling, and reducing the hidden dimension of the last linear layer to 100. For Split CIFAR-100, we use $\e = \frac{2}{255}$ during training.

The total number of learnable parameters used in each configuration is reported in Tab.~\ref{appendix:tab_learnable_params}. Furthermore, we adopt the same FGSM, PGD, and AutoAttack settings as described in the main paper.

\begin{table}[t]
\centering
\caption{Comparison of learnable parameters (in millions) between our method and the baseline from \cite{ru2024maintaining}.}
\label{appendix:tab_learnable_params}
\setlength{\tabcolsep}{4pt}
\footnotesize
\begin{tabular}{lccc}
\toprule
Method & Permuted MNIST & Rotated MNIST & Split CIFAR-100 \\
\midrule
Baseline & 2M & 2M & 5.5M \\
\our{}   & 2.8M & 2.8M & 5.6M \\
\bottomrule
\end{tabular}
\end{table}

Tables~\ref{appendix:tab_permuted_mnist}, \ref{appendix:tab_rotated_mnist}, and \ref{appendix:tab_split_cifar100} present the AA of \our{} and several baselines after learning all tasks sequentially. The results include performance under three adversarial attacks (AutoAttack, PGD, FGSM) and on original clean data.

On the Permuted MNIST benchmark (Tab.~\ref{appendix:tab_permuted_mnist}), \our{} delivers the strongest adversarial robustness among all compared methods. It surpasses prior approaches under all attack settings, including AutoAttack, PGD, and FGSM, establishing itself as the most resilient model to adversarial perturbations in this setting. While a few baselines, particularly DGP and GPM, achieve higher clean accuracy, their robustness drops notably under stronger attacks. In contrast, \our{} maintains both high robustness and solid performance on clean samples, without relying on overly large architectures or task-specific replay. This highlights the method’s ability to generalize under distribution shifts and adversarial pressure, a balance that remains a major challenge in continual learning.

\begin{table*}[ht]
\centering
\caption{Comparison of AA after completing all tasks on the Permuted MNIST dataset. AA results for \our{} are averaged over 2 seeds for AutoAttack and 5 seeds for all other evaluations.}
\begin{tabular}{lccccccc}
\hline
\multicolumn{1}{c}{\multirow{3}{*}{Method}} & \multicolumn{5}{c}{Permuted MNIST} \\ \cline{2-6} 
\multicolumn{1}{c}{} & AutoAttack & PGD & FGSM & \multicolumn{2}{c}{Original samples} \\ \cline{2-6} 
\multicolumn{1}{c}{} & AA(\%) & AA(\%) & AA(\%) & AA(\%) \\ \hline
SGD & 14.1 & 15.4 & 21.8 & 36.8 \\
SI & 14.3 & 16.5 & 22.3 & 36.9 \\
A-GEM & 14.1 & 19.7 & 22.9 & 48.4 \\
EWC & 39.4 & 43.1 & 50.0 & 84.9 \\
GEM & 12.1 & 75.5 & 72.8 & 96.4 \\
OGD & 19.7 & 24.1 & 26.0 & 46.8 \\
GPM & 70.4 &  72.9 & 65.7 & 97.2 \\
DGP & 81.6 & 81.2 & 75.8 & \textbf{97.6} \\ \hline
\our{} & \textbf{88.43} $\pm$ \textbf{0.35} & \textbf{90.47} $\pm$ \textbf{1.36} & \textbf{79.6} $\pm$ \textbf{1.34}  & 90.71 $\pm$ 1.8 \\ \hline
\end{tabular}
\label{appendix:tab_permuted_mnist}
\end{table*}

\begin{table*}[ht]
\centering
\caption{Comparison of AA after completing all tasks on the Rotated MNIST dataset. AA results for \our{} are averaged over 2 seeds for AutoAttack and 5 seeds for all other evaluations.}
\begin{tabular}{lccccccc}
\hline
\multicolumn{1}{c}{\multirow{3}{*}{Method}} & \multicolumn{5}{c}{Rotated MNIST} \\ \cline{2-6} 
\multicolumn{1}{c}{} & AutoAttack & PGD & FGSM & \multicolumn{2}{c}{Original samples} \\ \cline{2-6} 
\multicolumn{1}{c}{} & AA(\%) & AA(\%) & AA(\%) & AA(\%) \\ \hline
SGD & 14.1 & 9.9  & 20.4 & 32.3 \\
SI & 13.9  & 15.3  & 20.1 & 33.0 \\
A-GEM & 14.1  & 21.6  & 24.8 & 45.4 \\
EWC & 45.1  & 49.5  & 46.5  & 80.7  \\
GEM & 11.9  & 76.5  & 74.4  & 96.7  \\
OGD & 19.7  & 23.8  & 23.8  & 48.0  \\
GPM & 68.8  & 71.5  & 65.9  & 97.1  \\
DGP & 81.6 & 82.6 & 78.6 & $\textbf{98.1}$ \\ \hline
\our{} & \textbf{90.36} $\pm$ \textbf{0.1} & \textbf{92.76} $\pm$ \textbf{0.22} & \textbf{82.21} $\pm$ \textbf{0.43}  & $93.12 \pm 0.22$  \\ \hline
\end{tabular}
\label{appendix:tab_rotated_mnist}
\vskip 0.15in
\end{table*}

On Rotated MNIST (Tab.~\ref{appendix:tab_rotated_mnist}), \our{} sets a new state-of-the-art across all evaluated attacks. It achieves the highest Average Accuracy (AA) under AutoAttack (90.36\%), PGD (92.76\%), and FGSM (82.21\%), outperforming strong baselines like DGP and GPM by a significant margin. Moreover, \our{} not only maintains robust performance but also retains high AA (93.12\%) on original samples, closely matching the top-performing baselines in that category.

\begin{table*}[ht]
\caption{Comparison of AA after completing all tasks on the Split CIFAR-100 dataset. AA results for \our{} are averaged over 2 seeds for AutoAttack and 5 seeds for all other evaluations.}
\centering
\begin{tabular}{lcccccccc}
\hline
\multicolumn{1}{c}{\multirow{3}{*}{Method}} & \multicolumn{5}{c}{Split CIFAR-100} \\ \cline{2-6} 
\multicolumn{1}{c}{} & AutoAttack & PGD & FGSM & \multicolumn{2}{c}{Original samples} \\ \cline{2-6} 
\multicolumn{1}{c}{} & AA(\%) & AA(\%) & AA(\%) & AA(\%) \\ \hline
SGD & 10.3 & 12.8 & 19.4 & 46.5 \\
SI & 13.0 & 15.2 & 19.8 & 45.4 \\
A-GEM & 12.6 & 12.9 & 20.7 & 40.6 \\
EWC & 12.6 & 23.2 & 30.5 & 56.8 \\
GEM & 21.2 & 19.4 & 47.7 & 60.6 \\
OGD & 11.8 & 14.1 & 18.9 & 44.2 \\
GPM & 34.4 & 36.6 & \textbf{53.7} & 58.2 \\
DGP & 36.6 & 39.2 & 48.0 & \textbf{67.2} \\ \hline
\our{} & \textbf{54.55} $\pm$ \textbf{0.42} & \textbf{54.53} $\pm$ \textbf{0.48} & 47.71 $\pm$ 0.86 & 59.36 $\pm$ 1.11 \\ \hline
\end{tabular}
\label{appendix:tab_split_cifar100}
\vskip 0.15in
\end{table*}

On the more complex Split CIFAR-100 benchmark (Tab.~\ref{appendix:tab_split_cifar100}), \our{} once again delivers the strongest performance under adversarial attacks. It outperforms all baselines by a wide margin in both AutoAttack and PGD evaluations, achieving 54.55\% and 54.53\% AA, respectively, substantially ahead of the next best method, DGP. While some baselines achieve slightly better AA on original samples, they fall short under adversarial threat models. In contrast, \our{} consistently balances robustness and plasticity, confirming its capacity to learn durable representations without catastrophic forgetting, even in challenging continual learning settings.

While \our{} does not achieve the highest AA on original samples in this setup, it remains competitive with top methods like DGP. We emphasize that Interval MixUp was not used; our goal is to show that \our{} performs well even without it and with a smaller target network architecture. The slight drop in clean accuracy stems from two deliberate choices: (1) \our{} is trained using a relatively small target network architecture, and (2) training with stronger adversarial perturbations, which often lowers clean accuracy but improves robustness. This trade-off allows \our{} to strike a strong balance between robustness and performance, making it a reliable option for adversarially robust continual learning.

\section{Selected Hyperparameters}\label{sec:best_hyperparams}
For the Split CIFAR-100 experiments, we used an AlexNet-based target network along with a hypernetwork comprising two hidden layers of 100 and 50 neurons, respectively. The embedding size was set to 512, with a batch size of 32 and a learning rate of $0.001$. The hypernetwork used a regularization coefficient $\beta = 0.01$, and training was performed with an $\ell_\infty$ perturbation of $\e = 0.005$. The optimizer was Adam, and no data augmentation was used. Training was conducted for 200 epochs with ReLU activation in both the target and hypernetwork. We applied a learning rate scheduler: ReduceLROnPlateau with monitoring set to the maximum validation accuracy, reduction factor $\sqrt{0.1}$, patience of 5 epochs, minimum learning rate of $5\times10^{-7}$, and no cooldown. Batch normalization was enabled throughout. When Interval MixUp is applied, we sample the mixing coefficient from a Beta distribution with parameter $\alpha = 0.3$.

For the Permuted MNIST task, we employed a 2-layer MLP with 256 neurons per layer in the target network and 100 neurons per layer in the hypernetwork. The embedding size was 24, the batch size was 128, and the learning rate was 0.001. We used $\beta = 0.001$ and performed training with $\e = 0.01$. For the Rotated MNIST experiments, the setup mirrored that of Permuted MNIST but used an embedding size of 96. Training on Permuted MNIST and Rotated MNIST was conducted for 5000 iterations. When using Interval MixUp, we set the Beta distribution parameter to $\alpha = 0.1$.

In the Split miniImageNet experiments, we used a ResNet-18 target network and a hypernetwork with two hidden layers of 100 and 50 neurons, respectively. The embedding dimension was 128, with a batch size of 16 and a learning rate of $0.001$. We trained using the Adam optimizer for 50 epochs, applying ReLU activations and enabling batch normalization throughout. The hypernetwork was regularized with $\beta = 0.01$, and training included an $\ell_\infty$ perturbation of $\e = \frac{2}{255}$. No data augmentation was used. The learning rate scheduler matched that of the Split CIFAR-100 setup. When Interval MixUp was used, mixing coefficients were drawn from a Beta distribution with $\alpha = 0.2$.

Model selection was based on the best validation loss. The optimizer across all experiments was Adam, and ReLU was used as the activation function throughout. In all cases, the same adversarial attack settings were used as in \cite{ru2024maintaining}, except that the number of attack iterations for PGD was not specified in that work; we defaulted to 100 iterations for consistency across evaluations. For AutoAttack, we used the standard version with default parameters as suggested by the authors in \cite{croce2020reliableevaluationadversarialrobustness}. 

Experiments were conducted on NVIDIA RTX 4090 and A100 GPUs. All software and package versions are listed in the README file in our code repository.

\section{Training Details}\label{appendix:sec_training_details}
To identify the most effective configuration for each dataset, we conducted an extensive grid search over key hyperparameters, evaluating combinations to maximize validation performance. Below we summarize the search spaces used for each dataset.

\paragraph{Permuted MNIST.}
For the Permuted MNIST dataset, the grid search was performed using a 2-layer MLP with 256 neurons per layer in the target network. As a hypernetwork, we used an MLP, with architecture defined by the hypernetwork hidden layers parameter. The following hyperparameters were explored:
\begin{itemize}
\item Embedding sizes: 24, 48, 96
\item Learning rate: 0.001
\item Batch sizes: 64, 128
\item Regularization coefficients $\beta$: 0.0005, 0.001, 0.005, 0.01, 0.05
\item Perturbation sizes $\e$: $\frac{2}{255}$, $\frac{20}{255}$, $\frac{25}{255}$, 0.01
\item Hypernetwork hidden layers (MLP): [100, 50], [200, 50], [100, 100]
\item Number of training iterations: 5000
\item Number of validation samples per class in each task: 500
\item Beta distribution hyperparameters ($\alpha$): 0.01, 0.1, 0.2, 0.3, 0.4, 0.5
\end{itemize}

\paragraph{Rotated MNIST.}
We use the same grid search as for the Permuted MNIST dataset.

\paragraph{Split CIFAR-100.}
For the Split CIFAR-100 dataset, the grid search was conducted using AlexNet as the target network and an MLP-based hypernetwork. Batch normalization was enabled, and data augmentation was disabled. The full hyperparameter search space included:
\begin{itemize}
\item Embedding sizes: 128, 256, 512
\item Learning rate: 0.001
\item Batch sizes: 32, 64
\item Regularization coefficients $\beta$: 0.01, 0.05, 0.1
\item Perturbation sizes $\e$: $\frac{2}{255}$, $\frac{4}{255}$, 0.005, 0.01
\item Hypernetwork hidden layers (MLP): [200, 50], [100, 50], [100, 100], [100], [200]
\item Number of training epochs: 200
\item Number of validation samples per class in each task: 500
\item Beta distribution hyperparameters ($\alpha$): $\alpha$: 0.01, 0.1, 0.2, 0.3, 0.4, 0.5
\end{itemize}

\paragraph{Split miniImageNet.}
For the Split miniImageNet dataset, the grid search was conducted using ResNet-18 as the target network and an MLP-based hypernetwork. Batch normalization was enabled, and data augmentation was disabled. The full hyperparameter search space included:
\begin{itemize}
\item Embedding sizes: 96, 128, 256
\item Learning rate: 0.001
\item Batch size: 16
\item Regularization coefficients $\beta$: 0.01, 0.05
\item Perturbation sizes $\e$: $\frac{1}{255}$, $\frac{2}{255}$, $\frac{4}{255}$
\item Hypernetwork hidden layers (MLP): [200, 50], [100, 50], [100]
\item Number of training epochs: 50
\item Number of validation samples per class in each task: 250
\item Beta distribution hyperparameters ($\alpha$): $\alpha$: 0.01, 0.1, 0.2, 0.3, 0.4, 0.5
\end{itemize}

The Adam optimizer was used across all tasks. These grid searches enabled a systematic selection of optimal configurations for each dataset.

\paragraph{Interval Bound Propagation Training Details.}
Training with overly wide intervals can destabilize the learning process, especially in the early stages. To mitigate this, we begin training with $\kappa = 1$, placing full emphasis on the standard classification loss. As training progresses, the weight on the interval bound loss increases gradually, reaching $\kappa = 0.5$ by the midpoint of training. Simultaneously, we initialize the perturbation radius at $\e = 0$ and linearly increase it during the first half of training, so that it reaches the target value at the midpoint. Both $\kappa$ and $\e$ remain fixed for the remainder of training.

Formally, let $E$ be the total number of training iterations per task, and denote by $\kappa_i$ and $\e_i$ the values of $\kappa$ and $\e$ at iteration $i \in \{1, \ldots, E\}$. We define their schedules as:
\begin{align}
\kappa_i &= \max\{ \frac{1}{2},1 - \frac{i}{2E} \}
     \\
    \e_i &= \begin{cases}
        \frac{2i\cdot \e}{E} & \text{if } i \leq \lfloor\frac{E}{2}\rfloor,\\
        \e & \text{otherwise}.
    \end{cases}
\end{align}

\paragraph{Architecture Details.}
In the main part of the paper, we use the same base architecture across Permuted~MNIST, Rotated~MNIST, and Split-miniImageNet. For the Split~CIFAR-100, we adopt a streamlined variant of AlexNet to meet the capacity and efficiency requirements of hypernetwork-based training. Specifically, we remove dropout, include biases, and replace standard batch normalization layers with interval-aware variants. The classifier comprises two fully connected layers with 100 units each. To improve numerical stability during IBP, we replace max pooling with average pooling. Additionally, a $4 \times 4$ average pooling layer is applied after the final convolutional layer to reduce dimensionality.

This lightweight adaptation of AlexNet ensures that the target network remains compact, which is critical for training efficiency when its parameters are generated dynamically by a hypernetwork. Importantly, unlike~\cite{ru2024maintaining}, we do not allocate separate convolutional layers for each task; our model maintains a single shared architecture across tasks. Further architectural details and hyperparameter choices are discussed in Appendix Section~\ref{appendix:sec_training_details}. A shared MLP hypernetwork is used across all datasets.

\section{Ablation Study and Additional Experiments}\label{appendix:sec_ablation_study}

\subsection{Results on Permuted~MNIST}

On Permuted MNIST (Tab.~\ref{appendix:tab_permuted_mnist}), \our{} achieves competitive AutoAttack performance, closely matching DGP, while outperforming all baselines under PGD, FGSM, and on original samples. Notably, it is the only method with positive BWT, indicating both effective knowledge retention and improvement on previous tasks. \our{}\textsubscript{IM} further improves PGD and original accuracy, with a minor trade-off in BWT.

\begin{table}[!ht]
\centering
\caption{AA performance after all tasks on the Permuted~MNIST. Results for \our{} and \our{} with Interval MixUp (\our{}\textsubscript{IM}) are averaged over 2 seeds for AutoAttack and 5 seeds for the other evaluations. Baseline results are from \cite{ru2024maintaining}.}
\label{appendix:tab_permuted_mnist}
{\small
\begin{tabular}{@{}l@{}c@{}c@{}c@{}c@{}c@{}c@{}c@{}c@{}}
\hline
\multicolumn{1}{c}{Method} & \multicolumn{1}{c}{AutoAttack} & \multicolumn{1}{c}{PGD} & \multicolumn{1}{c}{FGSM} & \multicolumn{2}{c}{Original samples} \\ \hline 

\multicolumn{1}{c}{Metric} & AA(\%) & AA(\%) & AA(\%) & AA(\%) & BWT \\ 
\hline

\multicolumn{6}{c}{Permuted MNIST} \\
\hline
SGD & 14.1 & 15.4 & 21.8 & 36.8 & -0.66 \\
SI & 14.3 & 16.5 & 22.3 & 36.9 & -0.67 \\
A-GEM & 14.1 & 19.7 & 22.9 & 48.4 & -0.54 \\
EWC & 39.4 & 43.1 & 50.0 & 84.9 & -0.12 \\
GEM & 12.1 & 75.5 & 72.8 & 96.4 & -0.01 \\
OGD & 19.7 & 24.1 & 26.0 & 46.8 & -0.57 \\
GPM & 70.4 &  72.9 & 65.7 & 97.2 & -0.01 \\
DGP & \textbf{81.6} & 81.2 & 75.8 & 97.6 & -0.01 \\ \hline
\our{} & 80.91 & 90.11 & 78.87 & 93.58 & \textbf{0.02} \\ 
\our{}\textsubscript{IM}
 & 80.08 & \textbf{97.44} & \textbf{79.09} & \textbf{97.96} & -0.05 \\ \hline
\end{tabular}
}
\end{table}

\subsection{Alternative Epsilon Decay Rate Functions in Interval MixUp}\label{appendix:sec_decay_rates}
\begin{figure}[!h]
{\includegraphics[clip,width=0.98\linewidth]{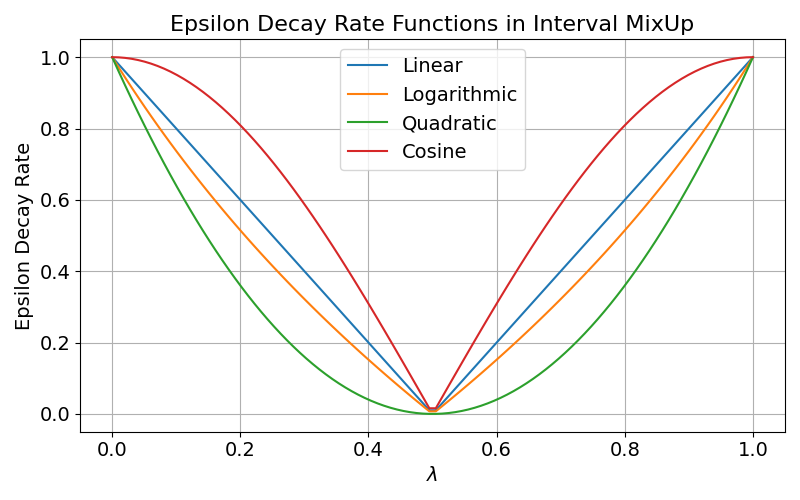} }
    \caption{Comparison of different epsilon decay rates used in Interval MixUp.}
\label{appendix:fig_decay_rates}
\end{figure}

In this subsection, we investigate how different epsilon decay rates affect the training of \our{}. In the paper, we employ a linear decay schedule. However, it is important to examine whether alternative decay strategies lead to different training behaviors. To this end, we additionally consider quadratic, logarithmic, and cosine decay functions. The corresponding plots of these decay schedules are shown in Figure \ref{appendix:fig_decay_rates}.

\begin{figure*}[t]
    \centering

    \begin{subfigure}[t]{0.24\textwidth}
        \centering
        \includegraphics[width=\linewidth]{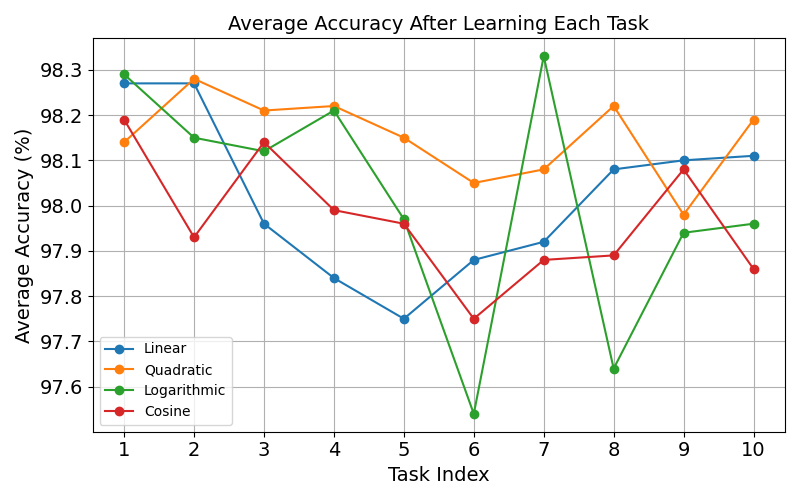}
        \caption{Permuted MNIST}
    \end{subfigure}
    \hspace{0.02\textwidth}
    \begin{subfigure}[t]{0.24\textwidth}
        \centering
        \includegraphics[width=\linewidth]{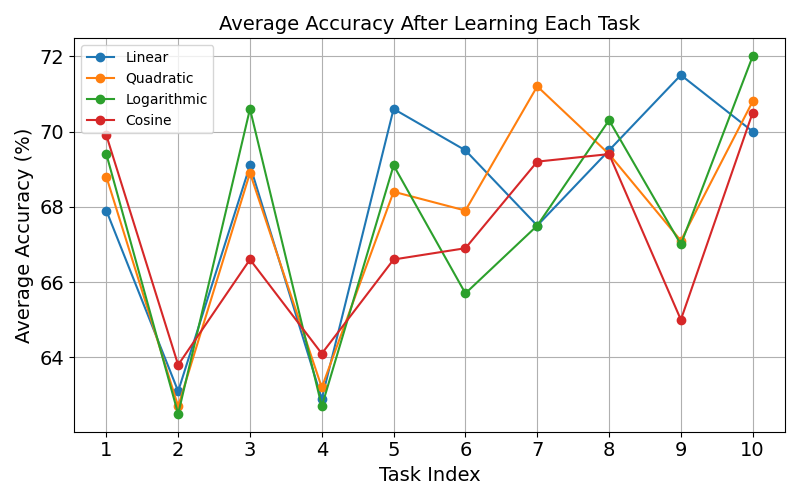}
        \caption{Split CIFAR-100}
    \end{subfigure}
    \hspace{0.02\textwidth}
    \begin{subfigure}[t]{0.24\textwidth}
        \centering
        \includegraphics[width=\linewidth]{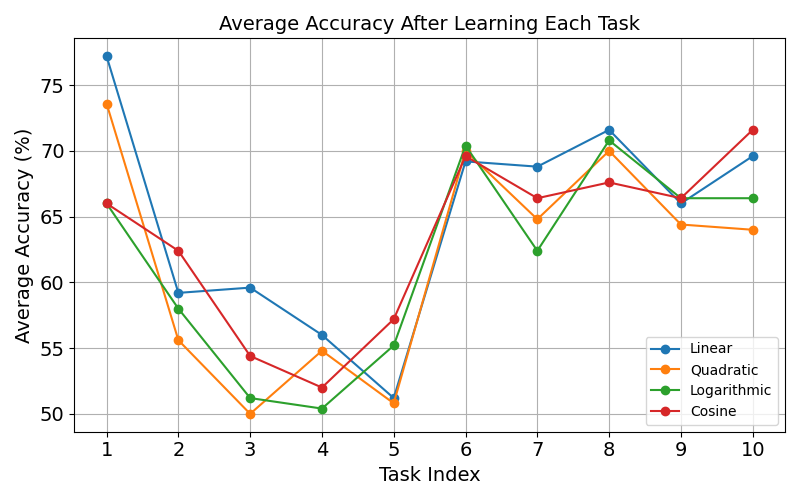}
        \caption{Split miniImageNet}
    \end{subfigure}

    \begin{subfigure}[t]{\textwidth}
        \centering
        \caption*{\textbf{AA after learning each task}}
    \end{subfigure}

    \begin{subfigure}[t]{0.24\textwidth}
        \centering
        \includegraphics[width=\linewidth]{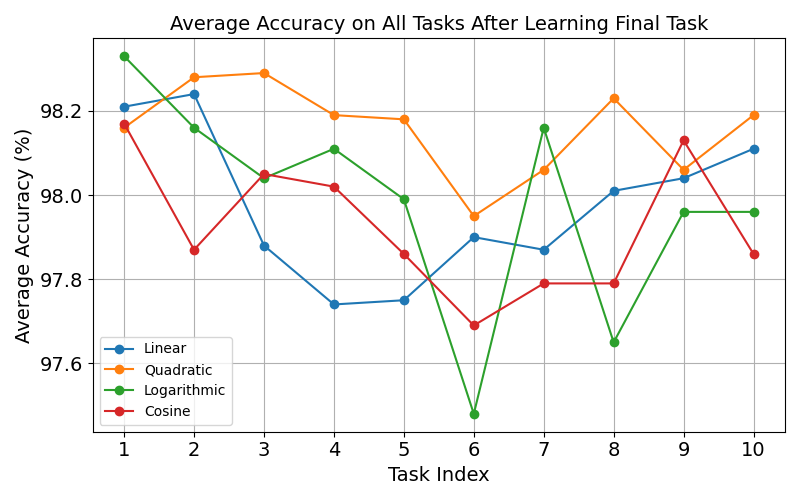}
        \caption{Permuted MNIST}
    \end{subfigure}
    \begin{subfigure}[t]{0.24\textwidth}
        \centering
        \includegraphics[width=\linewidth]{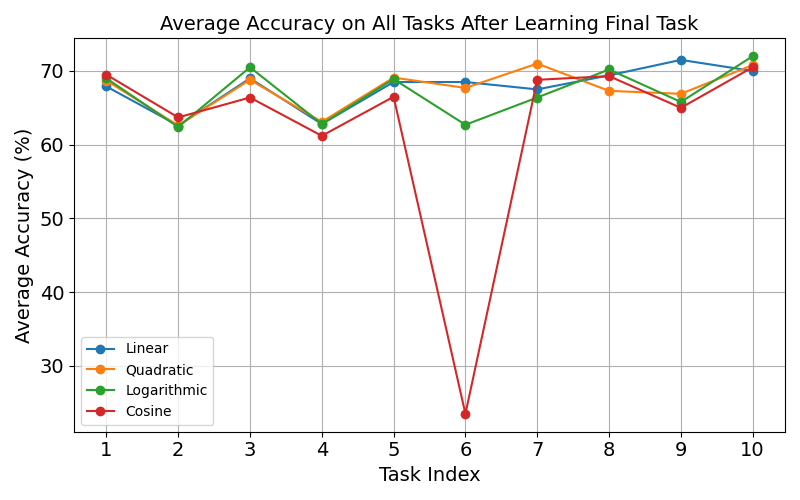}
        \caption{Split CIFAR-100}
    \end{subfigure}
    \begin{subfigure}[t]{0.24\textwidth}
        \centering
        \includegraphics[width=\linewidth]{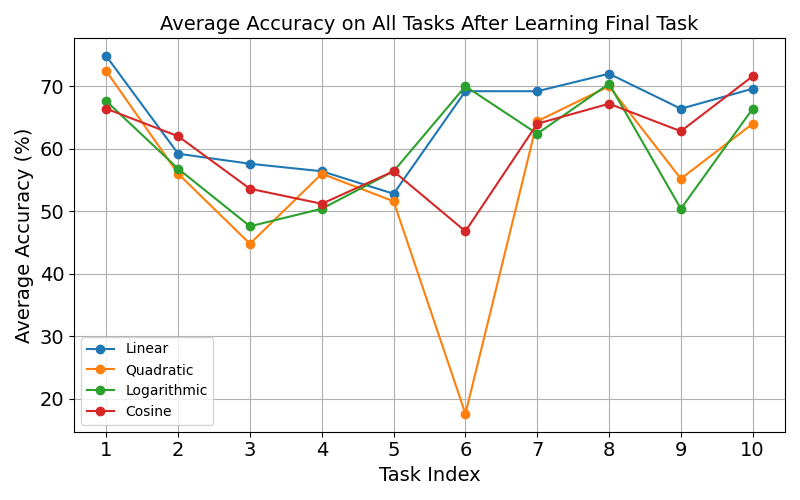}
        \caption{Split miniImageNet}
    \end{subfigure}
    \begin{subfigure}[t]{\textwidth}
        \centering
        \caption*{\textbf{AA on all tasks after training the final task}}
    \end{subfigure}

    \caption{Comparison of results for different epsilon decay rates across (a) Permuted MNIST, (b) Split CIFAR-100, and (c) Split miniImageNet. Top row: AA measured immediately after learning each task. Bottom row: AA evaluated on all tasks after completing the final task.}
\label{appendix:fig_epsilon_decay_rates_results}
\end{figure*}

Fig.~\ref{appendix:fig_epsilon_decay_rates_results} presents the AA over time for various decay rate strategies across three continual learning benchmarks: Permuted MNIST, Split CIFAR-100, and Split miniImageNet. The top row shows AA, computed as the mean accuracy over all tasks seen so far, immediately after learning each task. The bottom row shows final AA, calculated as the average accuracy on all tasks after the final task is learned.

Across all datasets, linear and logarithmic decay rate schedules demonstrate stable and competitive performance throughout training. In contrast, cosine and quadratic functions result in noticeably unstable and degraded accuracy, particularly evident in later tasks (such as the 6th task of Split CIFAR-100 and Split miniImageNet). This instability may be attributed to the more abrupt fluctuations of cosine and quadratic schedules (Figure \ref{appendix:fig_decay_rates}). These findings suggest that decay strategies, such as linear or logarithmic, better support robust learning in adversarial continual learning scenarios.

\begin{table}[ht]
\centering
\caption{Comparison of AA after completing all tasks on the Permuted MNIST dataset. AA results for different scheduling strategies (linear, quadratic, log, and cos) were calculated for a single seed.}
\label{appendix:tab_permuted_mnist_schedule}
{\small
\begin{tabular}{@{}l@{}c@{}c@{}c@{}c@{}c@{}}
\hline
\multicolumn{1}{c}{\multirow{3}{*}{Schedule}} & \multicolumn{5}{c}{Permuted MNIST} \\ \cline{2-6} 
\multicolumn{1}{c}{} & \multicolumn{1}{c}{AutoAttack} & \multicolumn{1}{c}{PGD} & \multicolumn{1}{c}{FGSM} & \multicolumn{2}{c}{Original samples} \\ \cline{2-6} 
\multicolumn{1}{c}{} & AA(\%) & AA(\%) & AA(\%) & AA(\%) & BWT \\ \hline
Linear & 80.53 & 97.48 & 79.4 & 98.01 & \textbf{0.002} \\
Quadratic & 80.45 & \textbf{97.64} & 79.51 & \textbf{98.16} & 0.001 \\
Log & 80.38 & 97.44 & 79.67 & 97.98 & -0.03 \\
Cos & \textbf{81.44} & 97.4 & \textbf{80.1} & 97.92 & -0.05 \\
\hline
\end{tabular}
}
\end{table}

\begin{table}[ht]
\centering
\caption{Comparison of AA after completing all tasks on the Split CIFAR-100 dataset. AA results for different scheduling strategies (linear, quadratic, log, and cos) were calculated for a single seed.}
\label{appendix:tab_split_cifar_100_schedule}
{\small
\begin{tabular}{@{}l@{}c@{}c@{}c@{}c@{}c@{}}
\hline
\multicolumn{1}{c}{\multirow{3}{*}{Schedule}} & \multicolumn{5}{c}{Split CIFAR-100} \\ \cline{2-6} 
\multicolumn{1}{c}{} & \multicolumn{1}{c}{AutoAttack} & \multicolumn{1}{c}{PGD} & \multicolumn{1}{c}{FGSM} & \multicolumn{2}{c}{Original samples} \\ \cline{2-6} 
\multicolumn{1}{c}{} & AA(\%) & AA(\%) & AA(\%) & AA(\%) & BWT \\ \hline
Linear & \textbf{63.89} & \textbf{62.9} & \textbf{46.99} &  66.97 & \textbf{-0.22} \\
Quadratic & 63.14 & 62.58 & 46.72 & \textbf{67.6} & -0.27 \\
Log & 62.91 & 62.1 & 46.75 & 67.07 & -0.68 \\
Cos & 58.87 & 57.72 & 42.91 & 62.44 & -5.29 \\
\hline
\end{tabular}
}
\end{table}

\begin{table}[ht]
\centering
\caption{Comparison of AA after completing all tasks on the Split miniImageNet dataset. AA results for different scheduling strategies (linear, quadratic, log, and cos) were calculated for a single seed.}
\label{appendix:tab_split_mini_imagenet_schedule}
{\small
\begin{tabular}{@{}l@{}c@{}c@{}c@{}c@{}c@{}}
\hline
\multicolumn{1}{c}{\multirow{3}{*}{Schedule}} & \multicolumn{5}{c}{Split miniImageNet} \\ \cline{2-6} 
\multicolumn{1}{c}{} & \multicolumn{1}{c}{AutoAttack} & \multicolumn{1}{c}{PGD} & \multicolumn{1}{c}{FGSM} & \multicolumn{2}{c}{Original samples} \\ \cline{2-6} 
\multicolumn{1}{c}{} & AA(\%) & AA(\%) & AA(\%) & AA(\%) & BWT \\ \hline
Linear & \textbf{60.08} & \textbf{60.32} & \textbf{56.16} & \textbf{64.72} & \textbf{-0.13} \\
Quadratic & 59.2 & 51.36 & 46.92 & 55.2 & -7.33 \\
Log & 55.72 & 55.64 & 51.64 & 59.84 & -2.09 \\
Cos & 55.52 & 55.84 & 51.88 & 60.2 & -3.51 \\
\hline
\end{tabular}
}
\end{table}

We evaluate the performance of \our{} trained using linear, quadratic, logarithmic, and cosine epsilon decay rate schedules in the context of adversarial attacks. The results are presented in Figures \ref{appendix:tab_permuted_mnist_schedule}, \ref{appendix:tab_split_cifar_100_schedule}, and \ref{appendix:tab_split_mini_imagenet_schedule}. The attacks used are the same as those described in the main paper. On Permuted MNIST, the results are nearly identical across all schedules. However, for Split CIFAR-100 and Split miniImageNet, the linear decay strategy performs the best. In this experiment, we used the best hyperparameters identified for the linear strategy in the main paper, so it is possible that further grid search for the other schedules could yield improved results.

\subsection{Impact of FGSM Perturbation Size on Average Accuracy}

\begin{figure*}[t!]
    \centering

    \begin{subfigure}[t]{0.32\linewidth}
        \includegraphics[width=\linewidth]{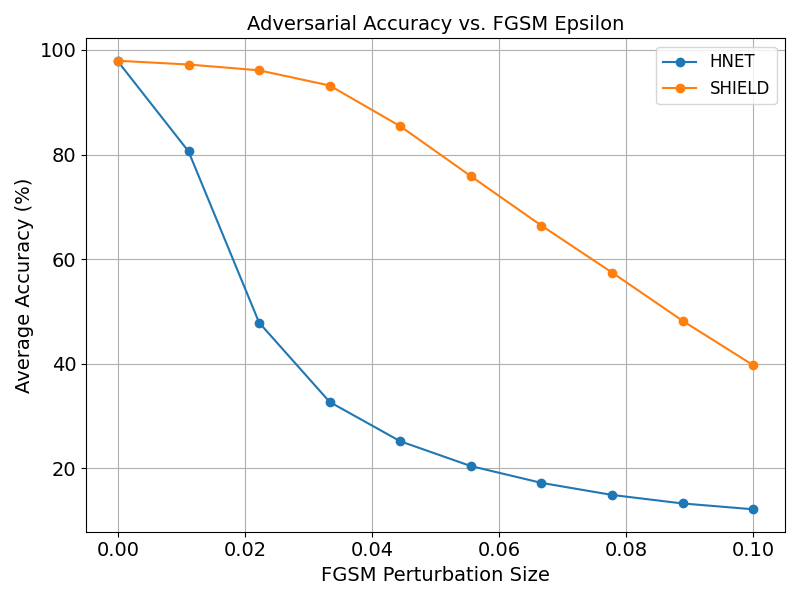}
        \caption{Permuted MNIST}
        \label{appendix:fig_permuted_mnist_fgsm}
    \end{subfigure}
    \begin{subfigure}[t]{0.32\linewidth}
        \includegraphics[width=\linewidth]{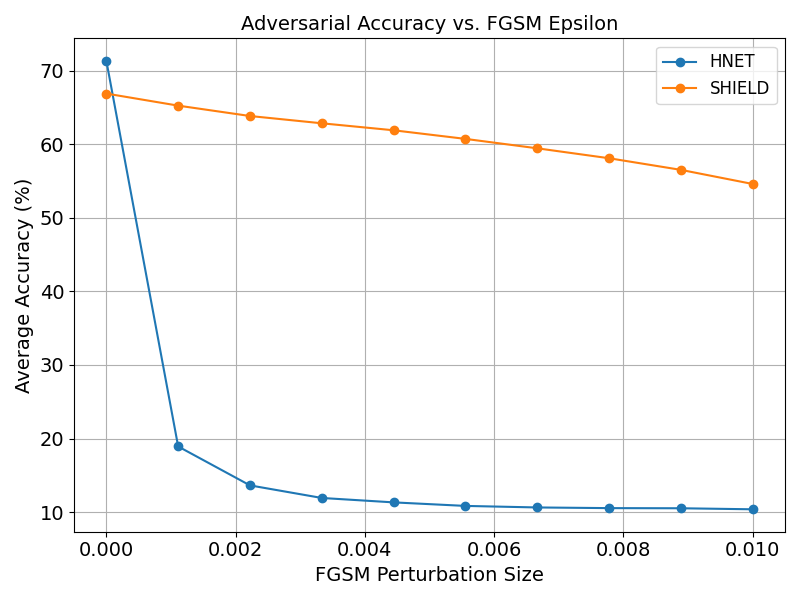}
        \caption{Split CIFAR-100}
        \label{appendix:fig_split_cifar_100_fgsm}
    \end{subfigure}
    \begin{subfigure}[t]{0.32\linewidth}
        \includegraphics[width=\linewidth]{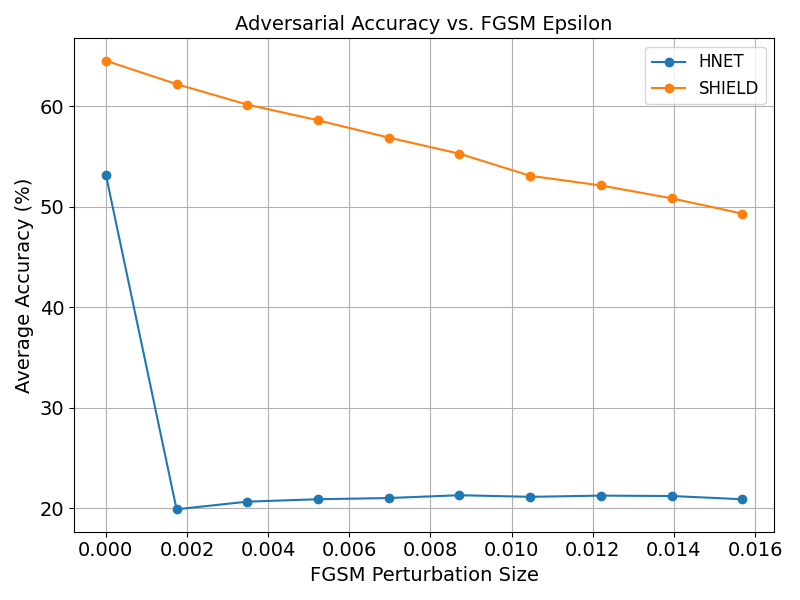}
        \caption{Split miniImageNet}
        \label{appendix:fig_split_mini_imagenet_fgsm}
    \end{subfigure}

    \caption{AA of \our{} and HNET under FGSM attacks with increasing perturbation sizes, evaluated on models trained on Permuted MNIST (Fig.~\ref{appendix:fig_permuted_mnist_fgsm}), Split CIFAR-100 (Fig.~\ref{appendix:fig_split_cifar_100_fgsm}), and Split miniImageNet (Fig.~\ref{appendix:fig_split_mini_imagenet_fgsm}).}

    \label{appendix:fig_fgsm_results}
\end{figure*}

In Fig.~\ref{appendix:fig_fgsm_results}, we show the effect of FGSM attacks with varying $\e_{\text{attack}}$ values on the AA metric for our method and for HNET. The results indicate that HNET is substantially more vulnerable to adversarial perturbations, with a steep drop in AA on Split~CIFAR-100 and Split~miniImageNet, even under small $\e_{\text{attack}}$ values. For each subfigure in Fig.~\ref{appendix:fig_fgsm_results}, we evaluate our models using 10 FGSM attack strengths, with $\e_{\text{attack}}$ values evenly spaced from 0 up to twice the $\e$ (perturbation value used in the training process) used during training. This setup allows us to assess how well the models generalize robustness beyond the training-time perturbation level. The optimal $\e$ values used are those identified in Section \ref{sec:best_hyperparams}.

\subsection{Impact of PGD Iterations on Average Accuracy}

\begin{figure*}[t!]
    \centering

    \begin{subfigure}[t]{0.32\linewidth}
        \includegraphics[width=\linewidth]{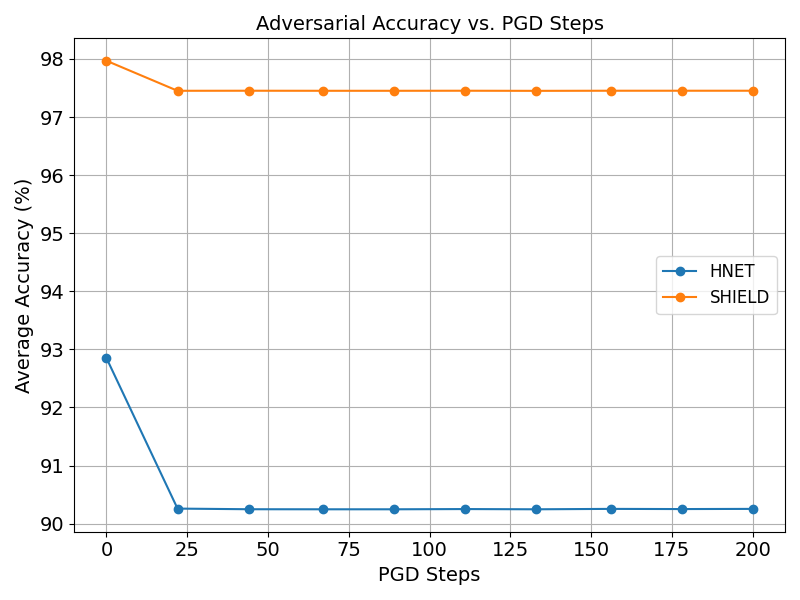}
        \caption{Permuted MNIST}
        \label{appendix:fig_permuted_mnist_pgd}
    \end{subfigure}
    \begin{subfigure}[t]{0.32\linewidth}
        \includegraphics[width=\linewidth]{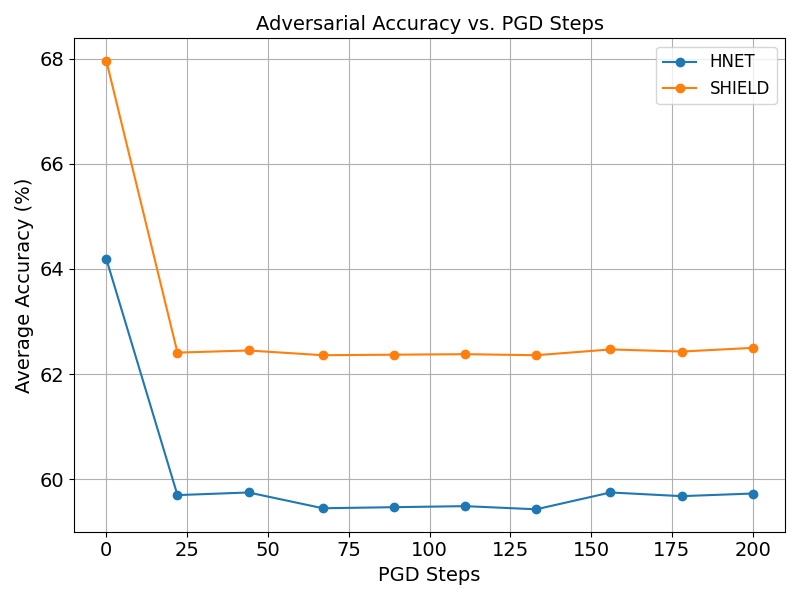}
        \caption{Split CIFAR-100}
        \label{appendix:fig_split_cifar_100_pgd}
    \end{subfigure}
    \begin{subfigure}[t]{0.32\linewidth}
        \includegraphics[width=\linewidth]{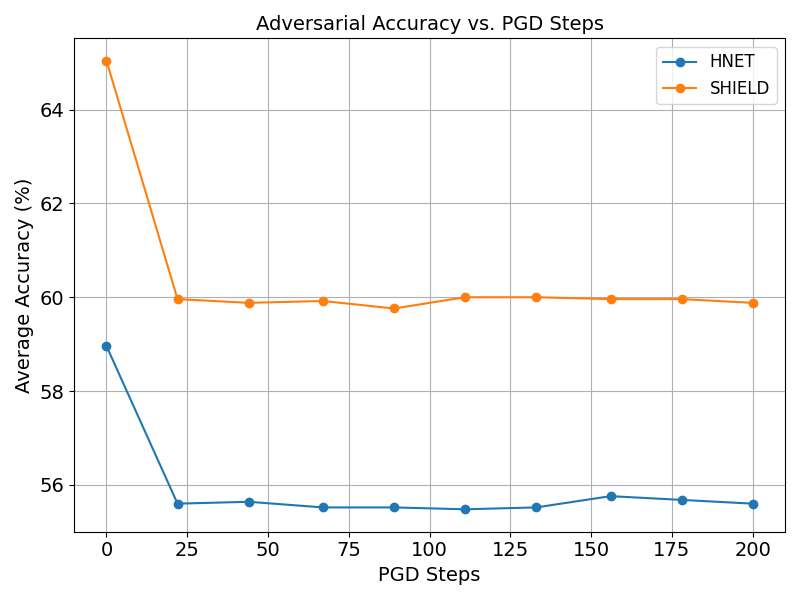}
        \caption{Split miniImageNet}
        \label{appendix:fig_split_mini_imagenet_pgd}
    \end{subfigure}

    \caption{AA of \our{} and HNET under PGD attacks with increasing number of gradient steps, evaluated on models trained on Permuted MNIST (Fig.~\ref{appendix:fig_permuted_mnist_pgd}), Split CIFAR-100 (Fig.~\ref{appendix:fig_split_cifar_100_pgd}), and Split miniImageNet (Fig.~\ref{appendix:fig_split_mini_imagenet_pgd}).}
\label{appendix:fig_pgd_results}
\end{figure*}

In Fig.~\ref{appendix:fig_pgd_results}, we present the effect of PGD attacks with an increasing number of steps on the AA metric for \our{} and HNET. Across all evaluated datasets, \our{} consistently achieves higher initial accuracy when subjected to no attack or only a few PGD steps. Both \our{} and HNET exhibit a noticeable drop in AA within the first 10 to 25 steps; however, their performance then stabilizes, demonstrating resilience to stronger iterative attacks.

For each subfigure in Fig.~\ref{appendix:fig_pgd_results}, we evaluate both models across 10 PGD configurations, with the number of attack steps evenly spaced from 0 to 200. The remaining PGD hyperparameters (e.g., step size, $\e_{\text{attack}}$) are kept identical to those used in the main experiments. This extended evaluation allows us to assess model robustness across a broader attack range than previously reported. As before, we use the best-performing hyperparameters identified in Section~\ref{sec:best_hyperparams}.

These results further confirm that \our{} provides stronger and more stable adversarial robustness, particularly under longer and more aggressive PGD attacks.

\subsection{Impact of $\beta$ Hyperparameter Selection on Training Process of \our{}}\label{appendix:exp_beta_param}

\begin{figure}[htbp]
    \centering
    \begin{subfigure}[b]{0.48\textwidth}
        \includegraphics[width=\textwidth]{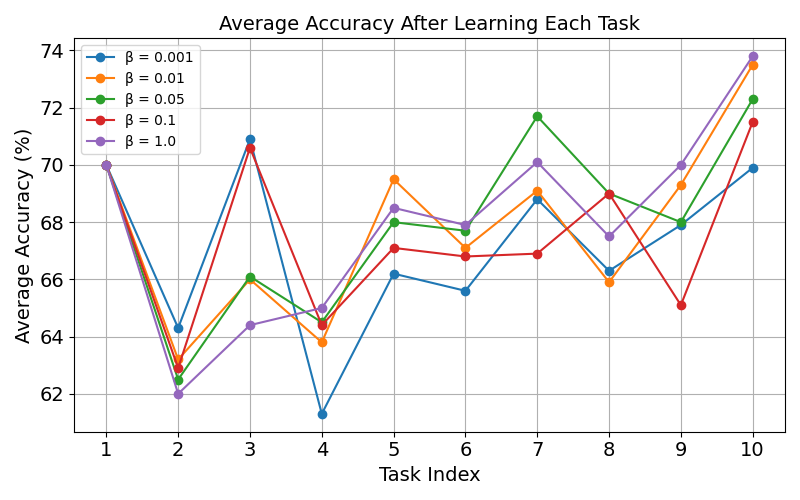}
        \caption{AA calculated directly after learning each task}
        \label{appendix:fig_beta_exp_after_each_task}
    \end{subfigure}
    \hfill
    \begin{subfigure}[b]{0.48\textwidth}
        \includegraphics[width=\textwidth]{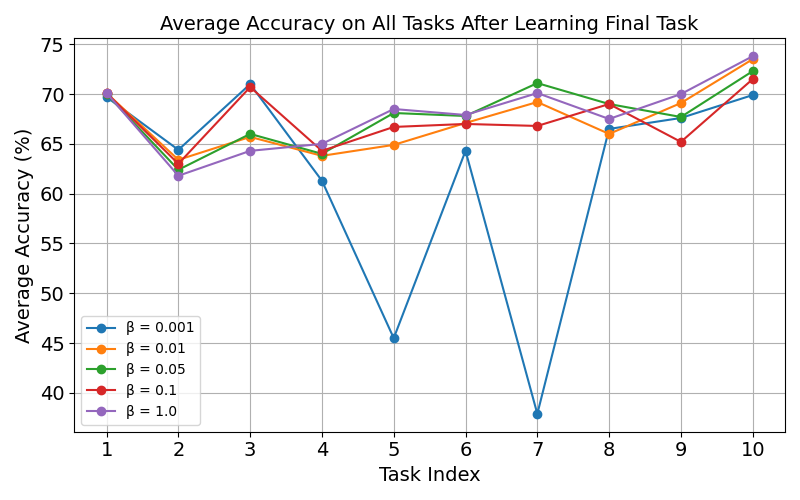}
        \caption{AA calculated after the final task}
        \label{appendix:fig_beta_after_final_task}
    \end{subfigure}
    \caption{Comparison of AA curves for varying $\beta$ values, which control hypernetwork regularization strength in \our{}}
    \label{appendix:fig_beta_accuracy_comparison}
\end{figure}

We evaluate the ability of \our{} to retain knowledge across tasks, which is controlled by the hyperparameter $\beta$. In Fig.~\ref{appendix:fig_beta_accuracy_comparison}, we present AA curves obtained by training the model with optimal hyperparameters while varying $\beta$. The results show that $\beta$ values of $0.01$ and $1.0$ yield the best performance, ensuring stable predictions and minimal forgetting. In contrast, a small value of $\beta = 0.001$ leads to significant forgetting, particularly on the 5th and 6th tasks, after learning the final task.

\subsection{Performance of \our{} on TinyImageNet dataset}
We assess the scalability of \our{} on a more challenging benchmark involving a larger number of tasks. Specifically, we use TinyImageNet, a subset of ImageNet, consisting of 200 classes, each with 500 training images and 50 validation images at a resolution of $64 \times 64$ pixels. To construct the benchmark, we define 40 tasks, each composed of 5 randomly selected, non-overlapping classes. This setup allows us to evaluate \our{}'s performance under increased task complexity and class diversity.

We compare the robustness of \our{} with the traditional HNET approach~\cite{von2019continual} under three adversarial attacks: FGSM, PGD, and AutoAttack. Both models are trained specifically on TinyImageNet. For FGSM, we set the attack strength to \(\e_{\text{attack}} = \frac{4}{255}\). For PGD and AutoAttack, we use \(\e_{\text{attack}} = \frac{2}{255}\), with a PGD step size of \(\delta = \frac{4}{255}\) and 100 iterations. The AutoAttack configuration is the same as used in the main experiments of the paper.

We use ResNet-18 as the target network. The hypernetwork is implemented as a multilayer perceptron (MLP) with a single hidden layer of 100 neurons and an embedding size of 48. Results are reported using a fixed random seed set to 1, shared across both models. We train both architectures for 50 epochs using the Adam optimizer, a learning rate of 0.001, and a batch size of 128. The same learning rate scheduler as in the Split CIFAR-100 experiments is applied. We set \(\beta = 0.05\). For \our{}, we use \(\e = 0.005\) during training and apply Interval MixUp.

\begin{table}[h]
\centering
\caption{Comparison of AA after completing all tasks on the TinyImageNet dataset.
}\label{appendix:tab_tinyimagenet}
{\small
\begin{tabular}{lcccc}
\toprule
Method & AutoAttack & PGD & FGSM & Original samples \\
\midrule
HNET & 46.37 & 45.3 & 49.17 & 63.83 \\
\our{} & \textbf{51.78} & \textbf{52.83} & \textbf{52.36} & \textbf{64.33} \\
\bottomrule
\end{tabular}}
\end{table}

From Tab.~\ref{appendix:tab_tinyimagenet}, we observe that even on a challenging dataset like TinyImageNet, our model achieves high AA on original (non-adversarial) test samples while also maintaining robustness against adversarial attacks. While HNET achieves competitive results, our method performs slightly better, especially under adversarial conditions.

\subsection{Impact of $\kappa$ Hyperparameter Selection on Average Accuracy}
To evaluate the sensitivity of our method to the hyperparameter $\kappa$, we conducted experiments on the Split CIFAR-100 dataset, measuring AA under various adversarial attacks (AutoAttack, PGD, FGSM) as well as on original samples, along with BWT. The results, averaged over 2 seeds, are presented in Tab.~\ref{appendix:tab_kappa_impact}. The table compares performance across $\kappa$ values of 0.0, 0.5, and 1.0, with the best results per metric highlighted in bold.
\begin{table}[ht]
\caption{Impact of the hyperparameter $\kappa$ on AA under various attacks and backward transfer on Split CIFAR-100. Results are averaged over 2 seeds. Bold values indicate the best performance per metric.}\label{appendix:tab_kappa_impact}
\centering
\begin{tabular}{lccccc}
\hline
\multirow{3}{*}{$\kappa$} & \multicolumn{5}{c}{Split CIFAR-100} \\ \cline{2-6}
 & AutoAttack & PGD & FGSM & \multicolumn{2}{c}{Original samples} \\ \cline{2-6}
 & AA(\%) & AA(\%) & AA(\%) & AA(\%) & BWT \\ \hline
0.0 & 61.24 & 60.1 & 45.01 & 64.93 & \textbf{0.06} \\ \hline
0.5 & \textbf{63.08} & \textbf{62.39} & \textbf{46.48} & \textbf{67.45} & -0.41 \\ \hline
1.0 & 56.58 & 58.94 & 45.54 & 64.55 & - 8.39 \\
\hline
\end{tabular}
\end{table}
From the results, $\kappa = 0.5$ consistently achieves the highest average accuracy across all evaluated attacks and on original samples, demonstrating superior robustness and generalization. While BWT is slightly negative for $\kappa = 0.5$, indicating minimal forgetting, it outperforms the other values, particularly $\kappa = 1.0$, which exhibits significant degradation in both accuracy and BWT. In contrast, $\kappa = 0.0$ shows moderate performance but falls short of the balanced improvements offered by $\kappa = 0.5$.
Based on these findings, we select $\kappa = 0.5$ for all experiments in the main paper, as it provides the optimal trade-off between adversarial robustness, clean accuracy, and knowledge retention in continual learning settings.

\subsection{Hypernetwork Ablation}

To assess the importance of the hypernetwork component in \our{}, we compare the hypernetwork with IBP training against a variant where the hypernetwork is replaced with EWC while keeping IBP training. In both settings, we additionally employ Interval MixUp to ensure a fair comparison. The backbone architectures (AlexNet/MLP, depending on the dataset) remain unchanged. Results are reported in Table~\ref{appendix:tab_hypenet_vs_ewc}.

\begin{table}[ht]
\centering
\caption{Ablation study comparing \our{} and EWC + IBP (both with Interval MixUp). Results report AA on original samples and are averaged over 5 seeds.}
\label{appendix:tab_hypenet_vs_ewc}
\footnotesize
\setlength{\tabcolsep}{8pt}
\renewcommand{\arraystretch}{1.2}
\begin{tabular}{lcc}
\toprule
Method & Permuted MNIST & Split CIFAR-100 \\
\midrule
EWC + IBP & $58.77 \pm 0.21$ & $15.97 \pm 0.04$ \\
\our{} & \textbf{97.96} $\pm$ \textbf{0.04} & \textbf{67.45} $\pm$ \textbf{0.28} \\
\bottomrule
\end{tabular}
\end{table}

Replacing the hypernetwork with EWC leads to a noticeable drop in performance, particularly on Split CIFAR-100. On Permuted MNIST, EWC with IBP achieves $58.77 \pm 0.21\%$ AA, while performance on Split CIFAR-100 decreases to $15.97 \pm 0.04\%$. Notably, AA on original samples in the EWC-based variant is already substantially lower, which further limits adversarial robustness under strong attacks. This suggests that the hypernetwork plays a crucial role in maintaining stable representations across tasks, which in turn supports stronger adversarial robustness.

Overall, the results indicate that the hypernetwork is an essential component of \our{}, contributing significantly to both continual knowledge retention and robustness under adversarial evaluation.

\subsection{IBP-Verified Robustness Curve on Permuted MNIST}

\begin{figure}[t!]
\centering
\includegraphics[width=0.65\linewidth]{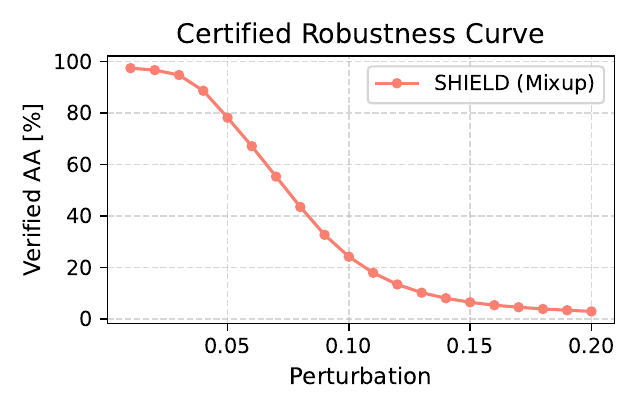}
\caption{IBP-verified AA as a function of the perturbation radius $\e$ for \our{} on Permuted MNIST, evaluated after the final task.}
\label{appendix:fig_certified_vs_eps_permuted_mnist}
\end{figure}

In Fig.~\ref{appendix:fig_certified_vs_eps_permuted_mnist}, we report the IBP-verified AA as a function of the perturbation radius $\e$. The curve is obtained using the model that achieved the best performance in the main paper and is evaluated after training on the final task. For each value of $\e$, a prediction is counted as correct only if the IBP bounds certify that the predicted class remains unchanged for all inputs within the corresponding $\ell_\infty$ neighborhood.

\section{Extended Results from Main Paper}\label{appendix:sec_extended_results}

In this section, we extend results from Table \ref{tab:results} to attach standard deviations. We present extended results in Table \ref{appendix:tab_permuted_mnist_extended}, \ref{appendix:tab_rotated_mnist_extended}, \ref{appendix:tab_split_cifar_100_extended}, and \ref{appendix:tab_split_mini_imagenet_extended}.

\begin{table*}[ht]
\caption{Comparisons of AA and BWT after learning all the tasks on Permuted~MNIST dataset.
The results for baselines were extracted from Table 1 in \cite{ru2024maintaining}.}
\label{appendix:tab_permuted_mnist_extended}
\centering
\begin{tabular}{lccccc}
\hline
\multirow{3}{*}{Method} & \multicolumn{5}{c}{Permuted MNIST} \\ \cline{2-6}
 & AutoAttack & PGD & FGSM & \multicolumn{2}{c}{Original samples} \\ \cline{2-6}
 & AA(\%) & AA(\%) & AA(\%) & AA(\%) & BWT \\ \hline
SGD & 14.1 & 15.4 & 21.8 & 36.8 & -0.66 \\
SI & 14.3 & 16.5 & 22.3 & 36.9 & -0.67 \\
A-GEM & 14.1 & 19.7 & 22.9 & 48.4 & -0.54 \\
EWC & 39.4 & 43.1 & 50.0 & 84.9 & -0.12 \\
GEM & 12.1 & 75.5 & 72.8 & 96.4 & -0.01 \\
OGD & 19.7 & 24.1 & 26.0 & 46.8 & -0.57 \\
GPM & 70.4 &  72.9 & 65.7 & 97.2 & -0.01 \\
DGP & \textbf{81.6} & 81.2 & 75.8 & 97.6 & -0.01 \\ \hline
\our{} & 80.91 $\pm$ 1.18 & 90.11 $\pm$ 0.8 & 78.87 $\pm$ 2.06 & 93.58 $\pm$ 0.52 & \textbf{0.02} $\pm$ \textbf{0.01} \\ 
\our{}\textsubscript{IM} & 80.08$\pm$0.01 & \textbf{97.44} $\pm$ \textbf{0.05} & \textbf{79.09} $\pm$ \textbf{0.87} & \textbf{97.96} $\pm$ \textbf{0.04} & -0.05 $\pm$ 0.03 \\

\hline
\end{tabular}
\end{table*}

\begin{table*}[ht]
\caption{Comparisons of AA and BWT after learning all the tasks on Rotated~MNIST dataset.
The results for baselines were extracted from Table 2 in \cite{ru2024maintaining}.}
\label{appendix:tab_rotated_mnist_extended}
\centering
\begin{tabular}{lccccc}
\hline
\multirow{3}{*}{Method} & \multicolumn{5}{c}{Rotated MNIST} \\ \cline{2-6}
 & AutoAttack & PGD & FGSM & \multicolumn{2}{c}{Original samples} \\ \cline{2-6}
 & AA(\%) & AA(\%) & AA(\%) & AA(\%) & BWT \\ \hline
SGD & 14.1 & 9.9  & 20.4 & 32.3 & -0.71 \\
SI & 13.9  & 15.3  & 20.1 & 33.0 & -0.72 \\
A-GEM & 14.1  & 21.6  & 24.8 & 45.4 & -0.57 \\
EWC & 45.1  & 49.5  & 46.5  & 80.7 & -0.18 \\
GEM & 11.9  & 76.5  & 74.4  & 96.7 & -0.01 \\
OGD & 19.7  & 23.8  & 23.8  & 48.0 & -0.55 \\
GPM & 68.8  & 71.5  & 65.9  & 97.1 & -0.01 \\
DGP & 81.6 & 82.6 & 78.6 & 98.1 & \textbf{-0.00} \\ \hline
\our{} & \textbf{85.64} $\pm$ \textbf{0.03} & 92.94 $\pm$ 0.16 &  \textbf{83.82} $\pm$ \textbf{0.2} & 95.62 $\pm$ 0.06 & -0.03 $\pm$ 0.05 \\
\our{}\textsubscript{IM} & 82.91 $\pm$ 0.37 & \textbf{97.88} $\pm$ \textbf{0.08} & 83.03 $\pm$ 0.45 & \textbf{98.32} $\pm$ \textbf{0.06} & -0.08 $\pm$ 0.04 \\
\hline
\end{tabular}
\end{table*}

\begin{table*}[ht]
\caption{Comparisons of AA and BWT after learning all the tasks on Split~CIFAR-100 dataset.
The results for baselines were extracted from Table 3 in \cite{ru2024maintaining}.}
\label{appendix:tab_split_cifar_100_extended}
\centering
\begin{threeparttable}
\begin{tabular}{lccccc}
\hline
\multirow{3}{*}{Method} & \multicolumn{5}{c}{Split CIFAR-100} \\ \cline{2-6}
 & AutoAttack & PGD & FGSM & \multicolumn{2}{c}{Original samples} \\ \cline{2-6}
 & AA(\%) & AA(\%) & AA(\%) & AA(\%) & BWT \\ \hline
SGD & 10.3 & 12.8 & 19.4 & 46.5 & -0.49 \\
SI & 13.0 & 15.2 & 19.8 & 45.4 & -0.48 \\
A-GEM & 12.6 & 12.9 & 20.7 & 40.6 & -0.48 \\
EWC & 12.6 & 23.2 & 30.5 & 56.8 & -0.35 \\
GEM & 21.2 & 19.4 & 47.7 & 60.6 & -0.13 \\
OGD & 11.8 & 14.1 & 18.9 & 44.2 & -0.50 \\
GPM & 34.4 & 36.6 & \textbf{53.7} & 58.2 & \textbf{-0.10} \\
DGP & 36.6 & 39.2 & 48.0 & 67.2 & -0.13 \\ \hline
\our{} & 60.91 & 59.77 & 45.37 & 64.24 & -0.34 \\ 
\our{}\textsubscript{IM} & \textbf{63.08 $\pm$ 0.64} & \textbf{62.39$\pm$ 0.38} & 46.48 $\pm$ 0.33 & \textbf{67.45 $\pm$ 0.28} & -0.41 $\pm$ 0.19 \\ \hline
\end{tabular}
\begin{tablenotes}
\item Note: In the original Table 3 from \cite{ru2024maintaining}, the reported accuracy for FGSM was higher than for original samples, which is inconsistent with expected adversarial behavior. We assume this was a mistake and have swapped the columns accordingly for a fair comparison.
\end{tablenotes}
\end{threeparttable}
\end{table*}

\begin{table*}[ht]
\caption{Comparisons of AA and BWT after learning all the tasks on Split~miniImageNet dataset.
The results for baselines were extracted from Figure 3 in \cite{ru2024maintaining}.}
\label{appendix:tab_split_mini_imagenet_extended}
\centering
\begin{tabular}{lccccc}
\hline
\multirow{3}{*}{Method} & \multicolumn{5}{c}{Split miniImageNet} \\ \cline{2-6}
 & AutoAttack & PGD & FGSM & \multicolumn{2}{c}{Original samples} \\ \cline{2-6}
 & AA(\%) & AA(\%) & AA(\%) & AA(\%) & BWT \\ \hline
SGD   & 20.5 & 22.0 & 23.5 & 30.8 & -0.24 \\
A-GEM & 19.0 & 19.8 & 21.2 & 29.2 & -0.28 \\
EWC   & 21.3 & 22.7 & 24.3 & 29.9 & -0.25 \\
SI    & 20.4 & 21.3 & 22.7 & 28.1 & -0.27 \\
GEM   & 22.3 & 23.8 & 25.4 & 31.8 & -0.20 \\
OGD   & 17.9 & 18.8 & 20.7 & 29.6 & -0.29 \\
GPM   & 26.3 & 27.1 & 28.8 & 36.8 & -0.12 \\
DGP   & 32.1 & 33.8 & 35.5 & 44.8 & \textbf{-0.05} \\\hline
\our{} & 56.22 $\pm$ 0.37 & 56.8 $\pm$ 0.95 & 53.08 $\pm$ 0.63 & 59.52 $\pm$ 0.67 & -0.16 $\pm$ 0.3 \\
\our{}\textsubscript{IM} & \textbf{57.9} $\pm$ \textbf{3.08} & \textbf{58.47} $\pm$ \textbf{1.9} & \textbf{54.1} $\pm$ \textbf{2.42} & \textbf{62.67} $\pm$ \textbf{1.9} & -0.18 $\pm$ 0.43 \\ \hline
\end{tabular}
\end{table*}

\section{Class-Incremental Learning Results}\label{appendix:sec_cil}

Tab.~\ref{appendix:tab_cil_adversarial_robustness} presents a comparison between \our{} and the HNET baseline in terms of adversarial robustness across four datasets: Permuted MNIST, Rotated MNIST, Split CIFAR-100, and Split miniImageNet, within the Class-Incremental Learning (CIL) setting. The results include AA under three adversarial attacks (AutoAttack, PGD, FGSM) as well as accuracy on original test samples, all reported after learning the full task sequence.

The results indicate that \our{} consistently achieves higher adversarial robustness than HNET under AutoAttack across all datasets. On Permuted MNIST and Rotated MNIST, \our{} shows large margins of improvement across all attack types, with particularly high AA under PGD on Permuted MNIST (97.21\%) and AutoAttack (36.9\%) on Rotated MNIST. For Split CIFAR-100, both methods perform similarly. On the more challenging Split miniImageNet benchmark, \our{} again significantly outperforms HNET in all scenarios, achieving over 12\% of AA under AutoAttack compared to just 2\% for the baseline.

These findings confirm that \our{} offers a substantial improvement in adversarial robustness in continual learning, even under challenging conditions. Notably, this is the first method, to our knowledge, that demonstrates strong robustness against adversarial attacks in the CIL setting, highlighting the practical effectiveness and novelty of the proposed framework.

The results reported in Tab.~\ref{appendix:tab_cil_adversarial_robustness} were averaged over multiple random seeds to ensure statistical reliability. Specifically, results for Permuted MNIST, Rotated MNIST, and Split CIFAR-100 were averaged over 5 seeds, while Split miniImageNet results were averaged over 3 seeds. Due to the computational cost of AutoAttack, its results were averaged over 2 seeds for all datasets.

\begin{table*}[ht]
\caption{
Adversarial robustness of \our{} across multiple datasets in the CIL scenario.
We report AA under AutoAttack, PGD, and FGSM attacks, as well as on original samples after learning all tasks.
}
\label{appendix:tab_cil_adversarial_robustness}
\centering
\begin{tabular}{lccccc}
\hline
Dataset & Method & AutoAttack & PGD & FGSM & Original samples \\
\hline
\multirow{2}{*}{Permuted MNIST}
    & \our{} & \textbf{79.76} $\pm$ \textbf{0.18} & \textbf{97.21} $\pm$ \textbf{0.32} & \textbf{77.76} $\pm$ \textbf{1.5} & \textbf{97.46} $\pm$ \textbf{0.27} \\
    & HNET   & 3.96 $\pm$ 2.45 & 89.16 $\pm$ 0.92 & 12.02 $\pm$ 0.26 & 93.97 $\pm$ 0.49 \\ \hline
\multirow{2}{*}{Rotated MNIST}
    & \our{} & \textbf{36.9} $\pm$ \textbf{0.21} & \textbf{39.12} $\pm$ \textbf{0.47} & \textbf{12.01} $\pm$ \textbf{0.28} & \textbf{42.32} $\pm$ \textbf{0.56} \\
    & HNET   & 7.77 $\pm$ 0.07 & 31.86 $\pm$ 0.75 & 3 $\pm$ 0.36 & 38.64 $\pm
    $ 0.8 \\ \hline
\multirow{2}{*}{Split CIFAR-100}
    & \our{} & \textbf{15.17} $\pm$ \textbf{0.23} & 13.08 $\pm$ 0.43 & 9.8 $\pm$ 0.47 & \textbf{23.11} $\pm$ \textbf{0.44} \\
    & HNET   & 10.57 $\pm$ 0.21 & \textbf{13.47} $\pm$ \textbf{0.5} & \textbf{11.91} $\pm$ \textbf{0.45} & 21.15 $\pm
    $ 0.69 \\ \hline
\multirow{2}{*}{Split miniImageNet}
    & \our{} & \textbf{12.66} $\pm$ \textbf{0.88} & \textbf{9.65} $\pm$ \textbf{1.79} & \textbf{9.53} $\pm$ \textbf{1.76} & \textbf{19.16} $\pm
    $ \textbf{1.25} \\
    & HNET   & 2.06 $\pm$ 0.42 & 4.33 $\pm$ 0.76 & 6.19 $\pm$ 0.92 & 12.84 $\pm$ 2.53 \\
\hline
\end{tabular}
\end{table*}

We present the pseudocode for \our{} in the CIL setting in Algorithm~\ref{appendix:alg_cil}. During inference, since task identity is unknown, the hypernetwork generates task-specific models for all tasks, and each is evaluated on the given input. The task is inferred by selecting the model with the lowest predictive entropy, assuming that the correct task model yields the most confident prediction. When adversarial attacks are considered, task inference is always performed on perturbed inputs to simulate realistic evaluation conditions. A prediction is counted as correct only if both the task is correctly inferred and the predicted class matches the true label; otherwise, it is treated as misclassified.

\begin{figure}[t]
\begin{algorithm}[H]
\caption{The pseudocode of \our{} in CIL setting}
\label{appendix:alg_cil}
\begin{algorithmic}[1]
\REQUIRE 
    Trained task embeddings $\{\boldsymbol{e}_t\}_{t=1}^T$, number of tasks $T$, trained hypernetwork $\mathcal{H}(\cdot; \boldsymbol{\Phi})$ with weights $\boldsymbol{\Phi}$, test sample $x$, number of classes $C_t$ per task

\ENSURE 
    Predicted label $\hat{y}$ and inferred task ID $\hat{t}$

\FOR{$t = 1$ to $T$}
    \STATE $\boldsymbol{\theta}_t \gets \mathcal{H}(\boldsymbol{e}_t; \boldsymbol{\Phi})$
    \STATE $\hat{y}^{(t)} \gets \text{Softmax}(f(x; \boldsymbol{\theta}_t))$
    \STATE Compute entropy: $\psi_t \gets -\sum_{j=1}^{C_t} \hat{y}^{(t)}_j \cdot \log(\hat{y}^{(t)}_j)$
\ENDFOR
\STATE $\hat{t} \gets \arg\min_t \psi_t$
\STATE $\hat{y} \gets \arg\max_{j} \hat{y}_j^{(\hat{t})}$
\end{algorithmic}
\end{algorithm}
\end{figure}

\section{Comparison with AIR Method}\label{appendix:sec_air_comparison}

Tab.~\ref{appendix:tab_cifar100_air_comparison} reports the performance of \our{} under the AIR evaluation protocol (None $\rightarrow$ FGSM $\rightarrow$ PGD) on CIFAR-100. \our{} achieves the highest accuracy on Task 2 and Task 3, outperforming all baselines under both FGSM and PGD attacks in the later tasks.

\begin{table}[htbp]
\centering
\caption{Accuracy results on CIFAR-100 for the 
None $\rightarrow$ FGSM $\rightarrow$ PGD training sequence. Results for \our{} are averaged over 3 random seeds. Baseline results are extracted from Tab.~4 in \cite{zhou2024defense}.}
\label{appendix:tab_cifar100_air_comparison}
\setlength{\tabcolsep}{3.5pt}
\footnotesize
\begin{tabular}{lccc}
\toprule
Method & Task 1 & Task 2 & Task 3 \\
\midrule
Vanilla            & 42.01 & 22.30 & 17.80 \\
EWC                & \textbf{48.35} & 22.54 & 16.59 \\
AIR                & 47.08 & 27.34 & 23.04 \\
\our{}             & 46.25 $\pm$ 0.29 & \textbf{28.36 $\pm$ 0.01} & \textbf{27.82 $\pm$ 0.08} \\
\midrule
Upper bound        & 45.33 & 30.23 & 21.25 \\
\bottomrule
\end{tabular}
\vspace{-0.3cm}
\end{table}

Importantly, \our{} is attack-agnostic - it does not rely on generating adversarial examples during training, in contrast to AIR. Instead, robustness is encouraged through interval-based training. For a fair comparison, we adopt the Wide ResNet architecture used in AIR, but employ a 10-layer variant with a widening factor of 10 to reduce the number of learnable parameters and ensure feasibility on our available hardware. Apart from this architectural adjustment, all experimental settings follow \cite{zhou2024defense}. Our hypernetwork consists of a two-layer MLP with 100 and 50 hidden units, respectively, and an embedding size of 512.

\section{Training Algorithm}\label{appendix:sec_train_algo}

The training procedure of \our{} is detailed in Algorithm~\ref{appendix:alg_training}. Note that subtracting a scalar from a tensor denotes a broadcasted operation, where the scalar is automatically expanded to match the shape of the tensor. This broadcasting allows element-wise arithmetic without the need for explicit reshaping. In particular, such operations occur during the construction of perturbed inputs (e.g., interval bounds), enabling efficient implementation.

\begin{figure}[t]
\begin{algorithm}[H]
\caption{The pseudocode for \our{} training}
\label{appendix:alg_training}
\begin{algorithmic}[1]
\REQUIRE 
    Task embeddings $\{\boldsymbol{e}_t\}_{t=1}^T$, hypernetwork $\mathcal{H}(\boldsymbol{e}_t; \boldsymbol{\Phi})$, target network $f(\cdot; \mathcal{H}(\boldsymbol{e}_t; \boldsymbol{\Phi}))$, number of tasks $T$, training data $D_t = \{(x_i, y_i)\}_{i=1}^{N_t}$, 
    regularization weight $\beta > 0$, 
    perturbation value $\e$, 
    cross-entropy weight $\kappa \in (0,1)$, 
    number of training steps $n$, 
    Beta distribution parameter $\alpha \geq 0$

\ENSURE 
    Trained parameters $\boldsymbol{\Phi}$ and $\{\boldsymbol{e}_t\}_{t=1}^T$

\STATE Randomly initialize $\boldsymbol{\Phi}$ and $\{\boldsymbol{e}_t\}_{t=1}^T$
\FOR{$t = 1$ to $T$}
    \IF{$t > 1$}
        \STATE Freeze $\boldsymbol{e}_{t-1}$
        \FOR{$j = 1$ to $t - 1$}
            \STATE $\boldsymbol{\theta}_j \gets \mathcal{H}(\boldsymbol{e}_j; \boldsymbol{\Phi})$
        \ENDFOR
    \ENDIF
    \STATE $\e' \gets 0$
    \STATE $\kappa' \gets 1$
    \FOR{$\text{step} = 1$ to $n$}
        \STATE $\boldsymbol{\theta}_t \gets \mathcal{H}(\boldsymbol{e}_t; \boldsymbol{\Phi})$
        \STATE Sample minibatch $\{(x_i, y_i)\}_{i=1}^B$ from $D_t$
        \STATE Sample $\lambda \sim \mathrm{Beta}(\alpha, \alpha)$
        \STATE $\e_{\text{IM}} \gets |2\lambda - 1| \cdot \e'$
        \STATE Sample $x_i \neq x_j$ from minibatch
        \STATE $\tilde{x} \gets \lambda \cdot x_i + (1 - \lambda) \cdot x_j$
        \STATE $\left[ \underline{\hat{y}}, \overline{\hat{y}} \right] \gets f([\tilde{x} - \e_{\text{IM}}, \tilde{x} + \e_{\text{IM}}]; \boldsymbol{\theta}_t)$
        \STATE $\hat{y} \gets f(\tilde{x}; \boldsymbol{\theta}_t)$
        \IF{$t = 1$}
            \STATE $\mathcal{L}_{\text{total}} \gets \mathcal{L}_{\text{IMixUp}}$ \COMMENT{From Eq.~\eqref{eq:interval_mixup_loss}, using $\e'$ and $\kappa'$}
        \ELSE
            \STATE $\mathcal{L}_{\text{total}} \gets \mathcal{L}_{\text{IMixUp}} + \beta \cdot \mathcal{L}_{\text{out}}$ \COMMENT{From Eq.~\eqref{eq:final_loss}, using $\e'$ and $\kappa'$}
        \ENDIF
        \STATE Update $\boldsymbol{\Phi}$ and $\boldsymbol{e}_t$ using gradient descent
        \IF{$\text{step} \leq \lfloor \frac{n}{2} \rfloor$}
            \STATE $\e' \gets \e \cdot \frac{2 \cdot \text{step}}{n}$
            \STATE $\kappa' \gets \max\{\frac{1}{2}, 1 - \frac{\text{step}}{2n}\}$
        \ENDIF
    \ENDFOR
    \STATE Store $\boldsymbol{e}_t$
\ENDFOR
\end{algorithmic}
\end{algorithm}
\end{figure}

\section{Training Time of \our{}}

Tab.~\ref{appendix:tab_trainingtime} reports the training time of \our{} model across different datasets. The durations are shown in the HH:MM:SS format and include standard deviations where available. For Permuted MNIST, Rotated MNIST, Split CIFAR-100, and Split miniImageNet, results are averaged over 5 training runs using the best-performing hyperparameters. For TinyImageNet, the time is reported from a single run with no standard deviation.

\begin{table}[ht]
\centering
\caption{Comparison of training time of the \our{} model on considered datasets.}
\label{appendix:tab_trainingtime}
\begin{tabular}{l c}
\toprule
Dataset & \our{} \\
\midrule
Permuted MNIST & $00:37:59 \pm 00:00:02$ \\
Rotated MNIST      & $00:36:18 \pm 00:00:19$ \\
Split CIFAR-100    & $02:10:23 \pm 00:02:00$ \\
Split miniImageNet & $09:21:05 \pm 00:12:11$ \\
TinyImageNet & $05:16:04$ \\
\bottomrule
\end{tabular}
\end{table}

\section{Interval MixUp Samples}\label{appendix:sec_interval_mixup_samples}
\begin{figure*}[!ht]
    \centering
    \includegraphics[width=\linewidth]{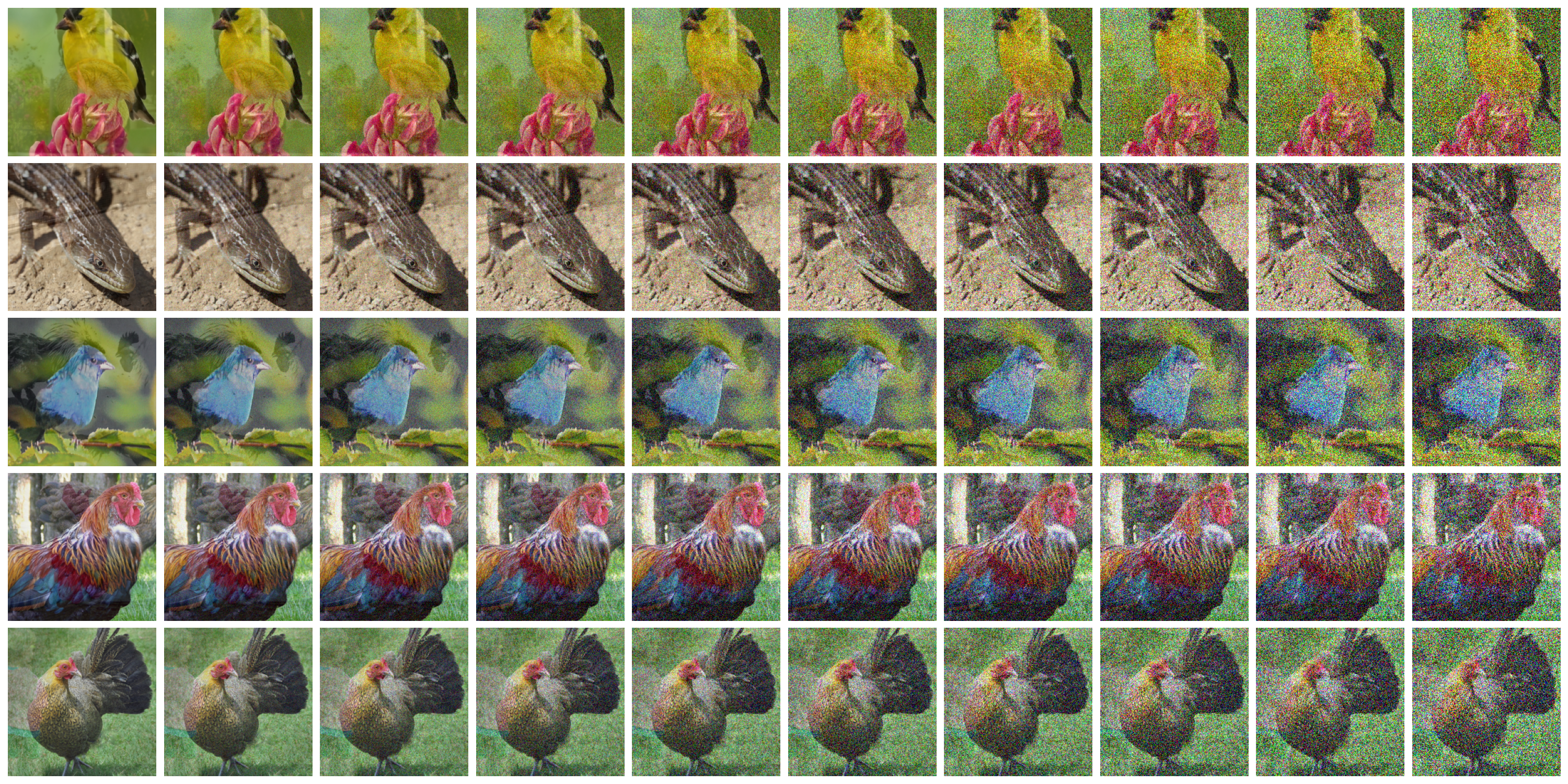}

    \caption{Each row corresponds to a single Interval MixUp-generated sample (hypercube center), and each column shows perturbations with increasing noise magnitude. This visualization demonstrates how added noise affects interpolated inputs, preserving semantic structure under small perturbations and gradually introducing variability.}
    \label{appendix:fig_interval_mixup_samples}
\end{figure*}

Fig.~\ref{appendix:fig_interval_mixup_samples} presents representative examples of samples generated using Interval MixUp. Each row corresponds to a single mixed input (the center of a hypercube), and each column shows perturbations with increasing noise magnitude. The samples were randomly selected from the Split~miniImageNet dataset. This visualization highlights how the neighborhood around each mixed point is explored, showing that small perturbations preserve the semantic content of the image, while larger ones introduce gradual variations.

\section{Theoretical Analysis of \our{}}\label{appendix:sec_theoretical_analysis}
We now present the theoretical analysis of \our{} together with the proof of the main theorem.
\subsection{Robustness Preservation Condition Theorem}

\begin{proof}
    Using the previously introduced notation, and following Definition~\ref{def:ibp_cert_robustness}, our goal is to show that for any task $s$ and any $(x, y_{\text{true}}) \in \mathcal{D}_s$, the following condition holds:
    \begin{equation}
        \underline{z}^{(K)}_{y_{\text{true}}}(x; \boldsymbol{\theta}_{s,t} + \boldsymbol{h}) > \max_{j \neq y_{\text{true}}} \overline{z}^{(K)}_j(x; \boldsymbol{\theta}_{s,t} + \boldsymbol{h}),
    \end{equation}
    where $\boldsymbol{h}$ denotes the hypernetwork output change induced by the gradient update of the hypernetwork weights while learning the $(t+1)$-th task. This parameter change affects the classifier outputs. For each classifier logit $l_x$, we can write:
    \begin{equation}
    \begin{aligned}
        f_{l_x}\left(x; \boldsymbol{\theta}_{s,t} + \boldsymbol{h}\right)
        \in \big[
        &\underline{z}^{(K)}_{l_x}(x; \boldsymbol{\theta}_{s,t})
           - \Delta_{\mathrm{max}}^{(t)}(\boldsymbol{h}), \\
        &\overline{z}^{(K)}_{l_x}(x; \boldsymbol{\theta}_{s,t})
           + \Delta_{\mathrm{max}}^{(t)}(\boldsymbol{h})
        \big].
    \end{aligned}
    \end{equation}

    Since each logit changes by at most $\Delta_{\mathrm{max}}^{(t)}(\boldsymbol{h})$, it follows that
    \begin{align}
        &\underline{z}^{(K)}_{y_{\text{true}}}(x; \boldsymbol{\theta}_{s,t} + \boldsymbol{h})
        - \max_{j \neq y_{\text{true}}} \overline{z}^{(K)}_{j}(x; \boldsymbol{\theta}_{s,t} + \boldsymbol{h})
        \\
        &\geq
        \underline{z}^{(K)}_{y_{\text{true}}}(x; \boldsymbol{\theta}_{s,t})
        - \max_{j \neq y_{\text{true}}} \overline{z}^{(K)}_{j}(x; \boldsymbol{\theta}_{s,t})
        - 2\,\Delta_{\mathrm{max}}^{(s,t)}(\boldsymbol{h}).
    \end{align}

    By the assumption stated in Eq.~\eqref{eq:robustness_assumption}, the right-hand side of the inequality above is strictly positive. Therefore, we obtain:
    \begin{equation}
        \underline{z}^{(K)}_{y_{\text{true}}}(x; \boldsymbol{\theta}_{s,t} + \boldsymbol{h})
        >
        \max_{j \neq y_{\text{true}}} \overline{z}^{(K)}_j(x; \boldsymbol{\theta}_{s,t} + \boldsymbol{h}),
    \end{equation}
    which concludes the proof.
\end{proof}

\subsection{Justification via Empirical Validation of the Theoretical Condition}

We now provide an empirical analysis to justify the practical efficacy of our method with respect to the theoretical conditions established in Theorem~\ref{thm:robustness_preservation}.

The theorem defines a strict, per-sample sufficient condition for preserving certified robustness: a sample $(x, y_{\text{true}})$ that is certifiably robust at time $t$ (i.e., $M(x; \boldsymbol{\theta}_{s,t}) > 0$) will remain so at time $t+1$ if its margin is large enough to absorb the logit change caused by the parameter update $\boldsymbol{h}$:
\begin{equation}
M(x; \boldsymbol{\theta}_{s,t}) > 2\,\Delta^{(s,t)}(x, \boldsymbol{h}),
\end{equation}
where $\Delta^{(s,t)}(x, \boldsymbol{h}) = \|f(x;\boldsymbol{\theta}_{s,t} + \boldsymbol{h}) - f(x;\boldsymbol{\theta}_{s,t})\|_{\infty}$.

\subsubsection*{The Practical Proxy Argument}

Explicitly enforcing this theoretical constraint for all samples at every step is computationally prohibitive in a data-free continual learning setting. We argue that the intrinsic regularization properties of our hypernetwork architecture serve as an effective, automatic, and computationally efficient proxy for this hard constraint.

Our hypernetwork regularization (see Eq.~\eqref{eq:hyper_loss_reg}) restricts changes in the hypernetwork weights ($\boldsymbol{\Phi}$). Since the output classifier weights ($\boldsymbol{\theta}_{s,t}$) are a smooth and continuous function of $\boldsymbol{\Phi}$, this regularization implicitly limits the magnitude of the resulting parameter update $\boldsymbol{h}$. Consequently, it constrains the logit change $\Delta^{(s,t)}(x, \boldsymbol{h})$.

\subsubsection*{Empirical Validation}

To validate this claim, Tab.~\ref{appendix:tab_assumption_check} reports the percentage of samples that satisfy the theoretical guarantee; i.e., samples that were initially robust (at time $t$) and for which the inequality $M(x; \boldsymbol{\theta}_{s,t}) - 2\,\Delta^{(s,t)}(x, \boldsymbol{h}) > 0$ holds after all tasks are completed. This directly measures how well our implicit regularization fulfills the conditions of the theorem.

The results are strong and consistent across all benchmarks. On Permuted MNIST, $97.70\%$ of samples satisfy the theoretical guarantee; on Rotated MNIST, $95.63\%$; and on Split CIFAR-100, $99.00\%$. These high rates indicate that our implicit regularization effectively limits the magnitude of the logit change $\Delta^{(s,t)}(x,\boldsymbol{h})$, such that, for the vast majority of inputs, this change is absorbed by the margin $M(x;\boldsymbol{\theta}_{s,t})$, thereby preserving the theoretical guarantee in practice.

\begin{table}[t]
\centering
\caption{Percentage of samples that remain robustly classified after learning the final task (Theorem~\ref{thm:robustness_preservation}). Higher values indicate better preservation of robustness. Results are averaged over 5 seeds.}
\label{appendix:tab_assumption_check}
\setlength{\tabcolsep}{3pt}
\footnotesize
\begin{tabular}{lccc}
\toprule
Task ID & Permuted MNIST & Rotated MNIST & Split CIFAR-100 \\
\midrule
1 & 97.67 $\pm$ 0.42 & 94.49 $\pm$ 0.53 & 100.00 $\pm$ 0.00 \\
2 & 97.09 $\pm$ 0.47 & 95.26 $\pm$ 0.87 & 95.21 $\pm$ 4.10 \\
3 & 96.83 $\pm$ 0.50 & 94.77 $\pm$ 1.07 & 100.00 $\pm$ 0.00 \\
4 & 97.38 $\pm$ 0.35 & 94.70 $\pm$ 1.53 & 99.04 $\pm$ 1.92 \\
5 & 97.23 $\pm$ 0.47 & 97.07 $\pm$ 0.78 & 99.00 $\pm$ 1.39 \\
6 & 98.02 $\pm$ 0.45 & 91.95 $\pm$ 4.03 & 98.95 $\pm$ 2.10 \\
7 & 97.84 $\pm$ 0.34 & 95.76 $\pm$ 0.63 & 99.00 $\pm$ 1.17 \\
8 & 98.55 $\pm$ 0.23 & 98.34 $\pm$ 0.14 & 99.76 $\pm$ 0.48 \\
9 & 98.73 $\pm$ 0.59 & 98.33 $\pm$ 0.22 & 100.00 $\pm$ 0.00 \\
\midrule
Average & 97.70 $\pm$ 0.11 & 95.63 $\pm$ 0.49 & 99.00 $\pm$ 0.59 \\
\bottomrule
\end{tabular}
\end{table}

\begin{figure*}[!ht]
    \centering
        \includegraphics[width=0.45\linewidth]{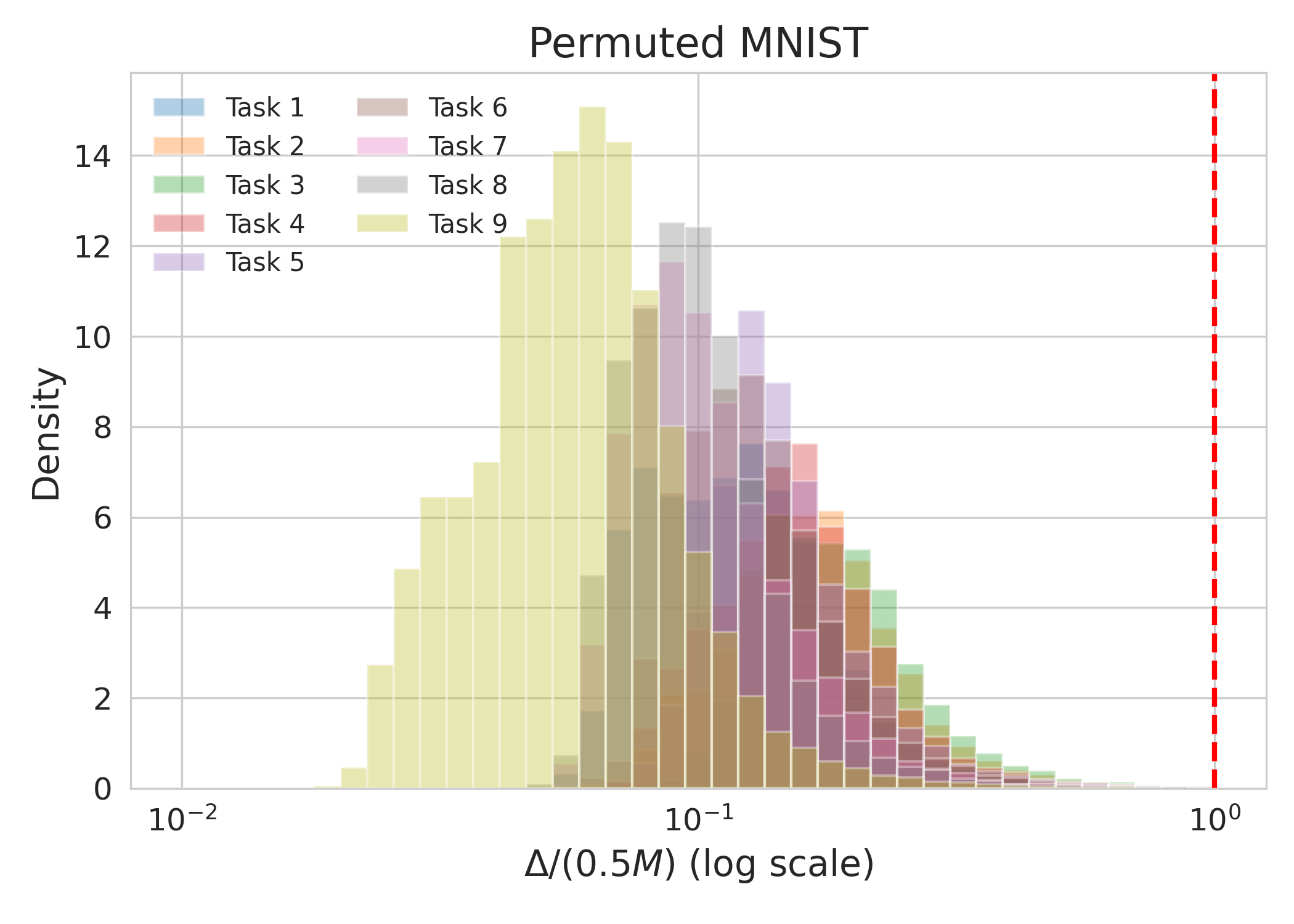}
        \includegraphics[width=0.45\linewidth]{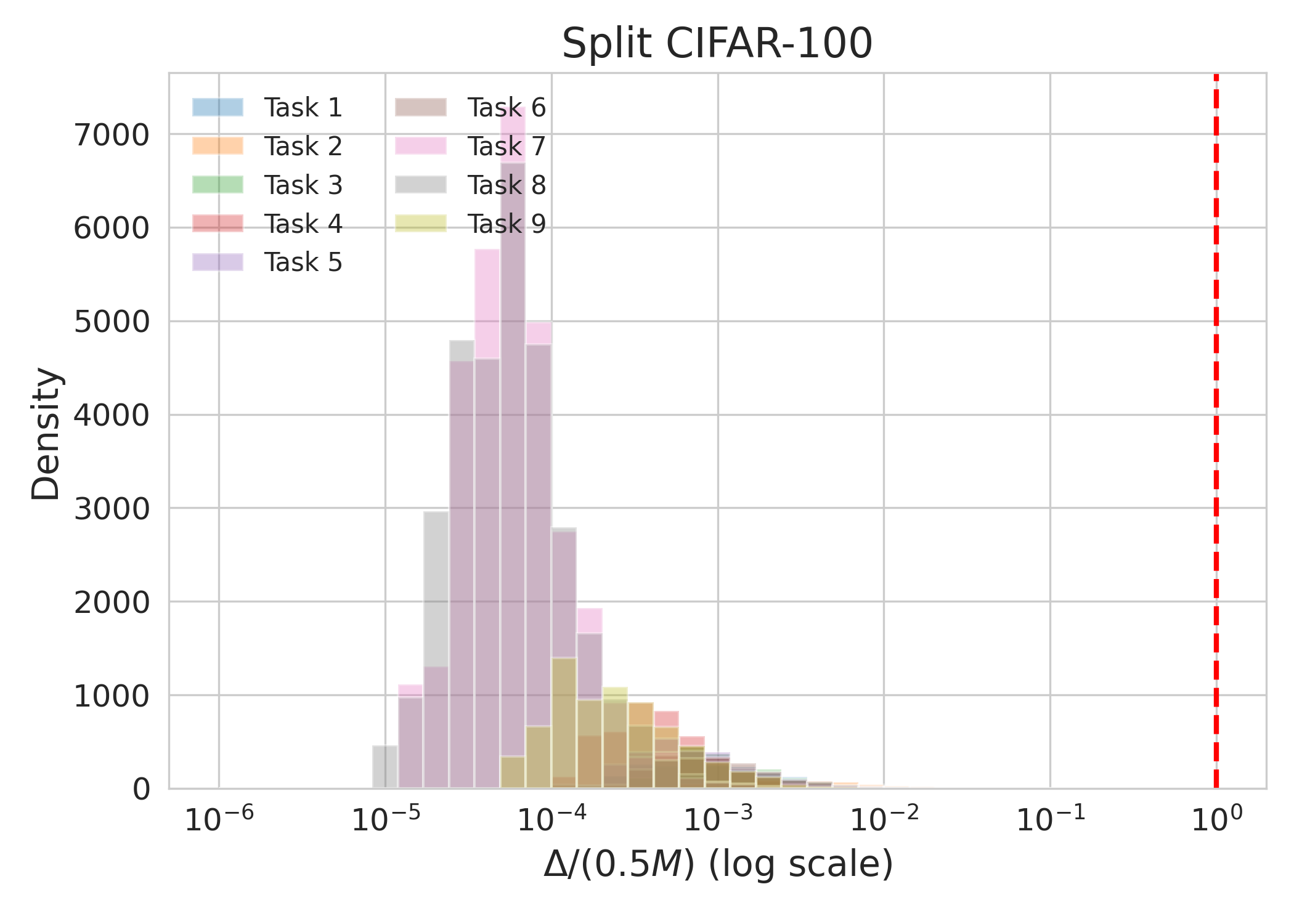}
   \caption{Empirical verification of Eq.~\eqref{eq:robustness_assumption} from Theorem~\ref{thm:robustness_preservation}. 
    The dashed red line marks the threshold $1$. 
    Left: Permuted MNIST. Right: Split CIFAR-100.}
    \label{appendix:fig_thm_assumption_check}
    \vspace{-0.4cm}
\end{figure*}

In addition, Fig.~\ref{appendix:fig_thm_assumption_check} provides a more detailed empirical verification of the underlying assumption. For each previous task $s$, we compute $\Delta_{\text{max}}^{(s,T)}(x;\boldsymbol{h})$, where $T$ denotes the final task and $\boldsymbol{h} = \boldsymbol{\theta}_{s,T} - \boldsymbol{\theta}_{s,s}$. Figure~\ref{appendix:fig_thm_assumption_check} shows histograms (log-scale on the $x$-axis) of the ratio 
$\Delta_{\text{max}}^{(s,T)}(x;\boldsymbol{h}) / \left(\tfrac{1}{2} M(x,y_{\text{true}};\boldsymbol{\theta}_{s,T})\right)$ 
for Permuted MNIST and Split CIFAR-100. The dashed red line indicates the critical threshold at $1$: values exceeding this threshold violate Eq.~\eqref{eq:robustness_assumption}, meaning that the parameter drift exceeds half of the certified margin and robustness is no longer guaranteed. Across both benchmarks, the mass of the distributions lies well below this threshold, indicating that the sufficient condition holds for nearly all IBP-certified samples, with a substantial margin. Therefore, robustness is preserved in practice.

In conclusion, our method maintains high certified robustness across tasks. The empirical analysis shows that knowledge of certified robustness is effectively transferred between tasks via the hypernetwork’s regularization mechanism. This provides a practical solution that satisfies the necessary condition for robustness preservation without incurring the computational cost of explicitly enforcing the hard theoretical constraint derived from Theorem~\ref{thm:robustness_preservation}.






\end{document}